\documentclass{article}

\usepackage{arxiv}

\usepackage[utf8]{inputenc} 
\usepackage[T1]{fontenc}    
\usepackage{hyperref}       
\usepackage{url}            
\usepackage{booktabs}       
\usepackage{amsfonts}       
\usepackage{nicefrac}       
\usepackage{microtype}      
\usepackage{graphicx}
\usepackage{doi}
\usepackage{subfig}
\usepackage{makecell}
\usepackage{multirow}

\title{An Enhanced Classification Method Based on Adaptive Multi-Scale Fusion for Long-tailed Multispectral Point Clouds}

\author{{LIU TianZhu}\\
	School of Electronics and Information Engineering\\
	Harbin Institute of Technology\\
	Harbin 150001, China\\
	\And
	{HU BangYan}\\
	School of Electronics and Information Engineering\\
	Harbin Institute of Technology\\
	Harbin 150001, China\\
	\And
	{GU YanFeng}\\
	School of Electronics and Information Engineering\\
	Harbin Institute of Technology\\
	Harbin 150001, China\\
	\And
	{LI Xian}\\
	School of Electronics and Information Engineering\\
	Harbin Institute of Technology\\
	Harbin 150001, China\\
    \texttt{xianli@hit.edu.cn} \\
	\And
	{Aleksandra Pi\v{z}urica}\\
	the Department of Telecommunications and Information Processing\\
	UGent-GAIM, Ghent University\\
	9000 Ghent, Belgium\\
}

\hypersetup{
pdftitle={An Enhanced Classification Method Based on Adaptive Multi-Scale Fusion for Long-tailed Multispectral Point Clouds},
pdfsubject={cs.CV},
pdfauthor={LIU TianZhu, HU BangYan, GU YanFeng, LI Xian, Aleksandra Pi\v{z}urica},
pdfkeywords={LiDAR, Multispectral Point Cloud, Classification, Multi-scale Fusion, Long-tailed Distribution},
}

\begin{document}
\maketitle

\begin{abstract}
Multispectral point cloud (MPC) captures 3D spatial-spectral information from the observed scene, which can be used for scene understanding and has a wide range of applications. However, most of the existing classification methods were extensively tested on indoor datasets, and when applied to outdoor datasets they still face problems including sparse labeled targets, differences in land-covers scales, and long-tailed distributions. To address the above issues, an enhanced classification method based on adaptive multi-scale fusion for MPCs with long-tailed distributions is proposed. In the training set generation stage, a grid-balanced sampling strategy is designed to reliably generate training samples from sparse labeled datasets. In the feature learning stage, a multi-scale feature fusion module is proposed to fuse shallow features of land-covers at different scales, addressing the issue of losing fine features due to scale variations in land-covers. In the classification stage, an adaptive hybrid loss module is devised to utilize multi-classification heads with adaptive weights to balance the learning ability of different classes, improving the classification performance of small classes due to various-scales and long-tailed distributions in land-covers. Experimental results on three MPC datasets demonstrate the effectiveness of the proposed method compared with the state-of-the-art methods.
\end{abstract}

\keywords{LiDAR \and Multispectral Point Cloud \and Classification \and Multi-scale Fusion \and Long-tailed Distribution}

\section{Introduction}
Multispectral point cloud (MPC) encompasses three dimensional (3D) spatial and spectral information, which can be used for the interpretation of land-cover composition and distribution from stereo observation perspective. It offers much more comprehensive scene analysis than the traditional stereo observation techniques, such as the single-band point cloud obtained by light detection and ranging (LiDAR) sensor that captures 3D spatial information of the land-covers but lacks spectral information, and multispectral imaging sensor, which is affected by the lighting conditions and lacks 3D spatial information. The emerging MPC Technology, enabling simultaneous acquisition of 3D spatial and spectral information from the observed scene, has a unique potential in a wide range of applications such as autonomous driving~\cite{1}, target detection~\cite{2}, segmentation~\cite{3,4}, and classification~\cite{5,6,53}.

MPCs can be generated in two ways: by an active or a passive method. Active method refers to acquiring data directly by multispectral LiDAR sensors. An example is Optech Titan system that utlizes three LiDAR sensors to acquire a three-band point cloud. This approach is straightforward but requires a precise alignment between point clouds in the different bands~\cite{7} and is typically confined to a smaller number of spectral bands. Passive MPC data~\cite{8,9} are reconstructed from multispectral images, after the steps of image preprocssing, correspondence computation, and 3D information recovery. An important advantage of this approach is that it generates MPCs without requiring special sensors, but there may be hollows in the generated MPC data.

In order to effectively utilize MPC, land-covers classification is an important downstream task. MPC classification can be classified into two classes: traditional methods and deep learning methods. Traditional classification methods fed the MPC into mathematically based classifiers such as support vector machines or random forests~\cite{10}. Early MPC classification methods directly utilized point-wise spectral information~\cite{11} and spectral index (e.g., NDVI) information~\cite{12,13,14}, and then methods that utilize both spectral and spatial information~\cite{15,16,17} are proposed. To further explore the classification methods, Dai et al.~\cite{18} explored first segmenting the scene and then refining the edge accuracy using spectral and spatial information; Ekhtari et al.~\cite{19} proposed to utilize point-wise spectral information with local spatial information for classification. In recent years, some methods~\cite{20,21} have explored the aggregation of multi-scale neighborhood space and spectral features. Juan et al.~\cite{22} explored a MPC acquisition system based on UAV and analyzed its classification effect using decision trees, extra trees, gradient boosting, random forest, and multilayer perceptron, respectively. In addition, some studies~\cite{23,24} have begun to investigate MPC classification methods for special scenes, such as fine classification of tree species. Although the above methods achieve a certain degree of MPC classification results, they are less effective when facing of more complex scenes because only shallow features in MPC can be extracted.

In recent years, deep learning methods have been gradually applied to land-cover classification tasks. MPC deep learning classification methods can be broadly categorized into three groups: projection methods, voxelization methods, and point processing methods. The projection method~\cite{25,26,27} projects the MPC into a multispectral image, then applies the multispectral image classification method to classify it, and finally projects the classification results back into the MPC. These methods, while more sophisticated, fail to capitalize on the unique advantages of MPC data. The voxelization method~\cite{28,29} is similar to the projection method in that the MPC is projected into the 3D mesh and the predictions are obtained and then re-projected into the MPC. Compared to the projection method, voxelization can better preserve the 3D spatial features in the MPC, but the corresponding computational effort also increases significantly when applied to large-scale point clouds.

For point processing methods, the first pioneering deep learning MPC classification method is PointNet~\cite{30}, which utilizes a multilayer perceptron to learn deep features from the land-covers, avoiding the problem that spatial convolution-based deep learning methods are difficult to apply to disordered MPC. Subsequently, a variety of MPC deep classification methods, such as PointNet++~\cite{31}, PointConv~\cite{32}, and Point Transformer~\cite{33}, are proposed. In 2020, Hu et al.~\cite{34} proposed a deep network RandLA-Net. Compared to methods such as PointNet, which uses the farthest point sampling method to downsampling the receptive field and aggregate deep featrures, RandLA-Net suggests that utilizing random sampling can improve the learning efficiency without degrading the accuracy, and drastically reduce the amount of computation, thus can be used for classification tasks in large-scale MPC datasets. Subsequently, various deep learning methods~\cite{35,36,37,38} based on the RandLA-Net framework are proposed, which added modules for aggregation of deep features to RandLA-Net to further improve the classification accuracy. In recent years, a variety of methods based on novel deep learning modules have been gradually proposed, exploring Transformer~\cite{39}, Feature Pyramid~\cite{40}, Gaussian Hypervector~\cite{41}, Capsule Transformer~\cite{42}, hypergraph~\cite{43}, multiple kernel learning~\cite{44}, etc., which have explored completely new ideas for MPC classification.

Although all of the above methods have achieved some success in MPC classification, most of the existing deep learning methods are tested on indoor datasets, and when the above methods are migrated to outdoor remote sensing datasets~\cite{45,46,47}, there are still several shortcomings as follows: First, compared to indoor datasets, the land-cover ground truths of outdoor datasets are difficult to obtain, so only sparsely labeled ground truths are generally available. Second, there are generally large scale differences in land-covers in remote sensing datasets, and it is difficult for a single receptive field to learn features applicable to all classes. In addition, there are large differences in the distribution of the number between different classes (i.e., the long-tailed distribution), which resulted in extremely low classification accuracy of some of the small classes, limiting the subsequent applications.

To tackle these challenges, a novel MPC classification method is proposed to enhance the classification accuracy under various-scale and long-tailed distribution datasets. Firstly, to solve the problem of sparsely labeled ground truths in MPC datasets, a grid-balanced sampling strategy is proposed to adaptively find the labeled truth values in the dataset and select the sample centroids according to the labeled ground truths rate, reliably extract training samples from a MPC with an unknown ground truths rate. Then, to improve the feature learning ability of the network for land-covers, a multi-scale feature fusion module is proposed to aggregate the deep spectral-spatial features output from different layers of the encoder and feed them into the decoder layer to improve the preservation ability of multi-scale detail features of the land-covers. Finally, to solve the problem of scale differences and long-tailed distribution of land-covers in outdoor datasets, an adaptive hybrid loss module are developed, which can balance the learning ability of each category and improve the overall classification accuracy. The main contributions of this article are summarized as follows.

(1) A classification framework based on the enhancement of adaptive multi-scale fusion for MPCs with long-tailed distributions is proposed, which improves all the inference processes of training sample generation, feature learning, and classification through the data characteristic of MPC. The main advantage of the proposed method is its feature extraction capability, exhibited as a remarkably improved performance when applied to challenging datasets.

(2) A multi-scale feature fusion module is devised to enhance the feature representation capacity. Compared with the traditional skip connection layer, this method aggregates the fine land-cover features obtained from different receptive fields in the encoder and feeds them into the decoder to improve the retention ability of the network for fine features at different scales.

(3) An adaptive hybrid loss module is developed to utilize adaptive weights with multiple classification heads, enhancing the feature learning capacity for various-scale and small classes.

The rest of this article is organized as follows. Section 2 describes the proposed MPC classification method in detail. The experimental results and ablation study analysis are reported in Section 3. Finally, the conclusion is drawn in Section 4.

\section{Methodology}
\subsection{Overall architecture}
\begin{figure}[!t]
    \centering
    \includegraphics[width=0.98\linewidth]{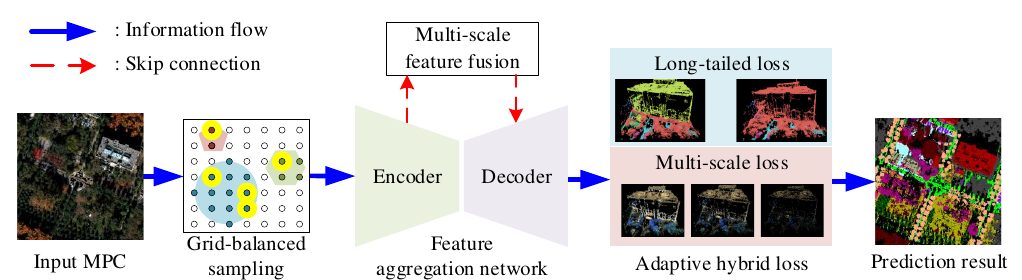}
    \caption{The overall architecture of proposed network. It comprises a grid-balanced sampling strategy, a feature aggregation network with multi-scale feature fusion module, an adaptive hybrid loss module. MPC repersents multispectral point cloud, blue solid arrows illustrate information flow and red dashed arrows represent skip connections.}
    \label{fig1}
\end{figure}

The workflow of the proposed method can be seen in Figure~\ref{fig1}. For the input MPC, training and test samples ${\mathbf{X}} \in {\mathbb{R} ^{N \times \left( {3 + d} \right)}}$ are obtained through a grid-balanced sampling strategy, where $N$ denotes the size of the samples, and $d$ is the number of spectral bands. A feature aggregation network is applied to extract point-wise deep spectral-spatial features ${\mathbf{F}} \in {\mathbb{R} ^{N \times C}}$ from the input samples, where $C$ is the number of features. Ultimately, the obtained deep spectral-spatial features ${\mathbf{F}}$ are fed into the multi-classification head and the feature extraction network is optimized by an enhanced adaptive fusion loss module.

Inspired by the lightweight and highly-efficient MPC classification network RandLA-Net~\cite{34}, the proposed feature aggregation network is based on a typical encoder-decoder architecture, with five layers each. In the encoder layer, spatial and spectral features are aggregated through a simple feature encoding method, and a random sampling strategy is used to downsample input features to improve the inference speed. The point-wise deep spectral-spatial features are restored using a simple MLP layer in the decoder layer. Between the encoder and the decoder layers, a multi-scale feature fusion module is used as a skip-connection to enhance the feature aggregation network's ability to retain fine features of the land-covers.

\subsection{Grid-balanced sampling}
In outdoor remote sensing datasets, manual labeling of land-cover is rather expensive and the acquired ground truth data are typically sparsely spatially labeled (as shown in Figure 6-8). The proportion of labeled points in the training samples obtained by traditional random sampling cannot be guaranteed, resulting in reduced network robustness. In addition, there is inevitably an overlapping area between the training samples and the test samples obtained by random sampling, which reduces the credibility of the evaluation of network classification accuracy. Therefore, a grid-balanced sampling strategy is proposed to extract training and test samples based on the spatial distribution of sparse ground truth adaptively. Compared with the traditional random sampling strategy, the proposed strategy ensures that the acquired training samples contain enough ground truth labels with the separation of training and test samples, which improves the robustness and credibility of the proposed method.

\begin{figure}[!t]
    \centering
    \includegraphics[width=0.68\linewidth]{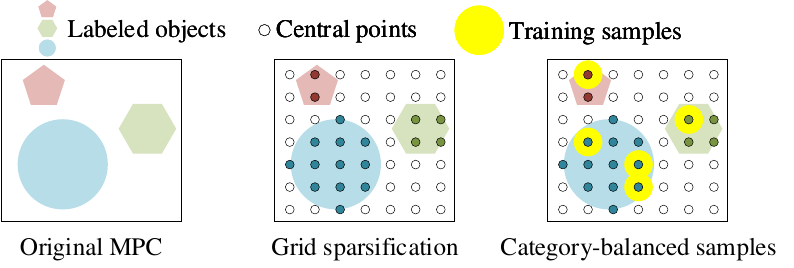}
    \caption{Illustration of the proposed grid-balanced sampling strategy. The colors show the category of points, yellow regions denotes training samples, others are test samples.}
    \label{fig2}
\end{figure}

The proposed grid-balanced sampling strategy consists of three steps: grid sparsification, category-balanced sampling, and sample generation. First, an appropriate downsampling ratio is chosen according to the size of the input receptive field. The purpose of this step is to divide the original MPC into non-overlapping grid areas to facilitate the selection of training samples and test samples; Then the sparse downsampled MPC is generated as centroids for input samples. The label of each centroids is the label of the highest-proportioned category of points in the corresponding area (including unlabeled points). Subsequently, the number of points in each category in the sparse MPC is calculated, and a certain proportion (5\% in this article), of each category is randomly chosen as the centroids for the training samples. Note that unlabeled centroids are not selected in this step, avoiding the selection of samples where most of the area are unlabeled points. Finally, the $k$-NN method is used to select the training samples around centroids from the original MPC. The remaining centroids are used to generate test samples. The above process can ensure that the labeled points account for the majority of the training samples, thereby improving the training efficiency. The proposed grid-balanced sampling strategy is shown in Figure~\ref{fig2}.

\subsection{Multi-scale feature fusion}
Due to the significant differences in the scales of land-covers in remote sensing datasets, it is difficult to extract fine features at a single scale for all land-cover types, leading to variations in the effectiveness of feature aggregation networks to learn classification features across different scales. To mitigate this issue, we propose a multi-scale feature fusion (MSFF) skip connection module. The idea behind this is that skip connections can feed fine features from the encoder layers into the decoder layers, reducing the depth degradation of the network, and enhancing the final classification results of aggregated point-wise deep features.

\begin{figure}[!t]
    \centering
    \includegraphics[width=0.98\linewidth]{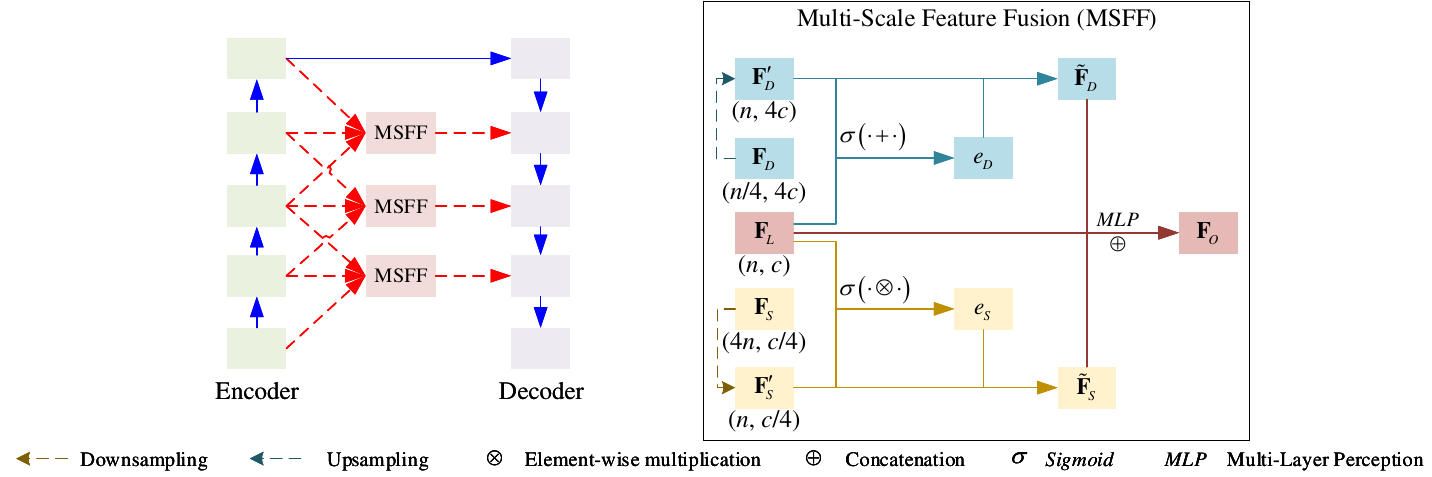}
    \caption{Architecture of the multi-scale feature fusion skip connection module. Left: the structure of the feature aggregation network. Right: proposed multi-scale feature fusion skip connection module. ${{\mathbf{F}}_S}$ represents shallow features, ${{\mathbf{F}}_L}$ represents local features, ${{\mathbf{F}}_D}$ represents deep features, ${{\mathbf{F}}_O}$ represents output features.}
    \label{fig3}
\end{figure}

The proposed MSFF method aggregates effectively fine features in the neighboring receptive fields, and improves this way significantly classification accuracy when facing variations in land-cover scales. Since the proposed encoder-decoder network has five layers, three MSFF modules are used. The structure of the MSFF module is shown in Figure~\ref{fig3}. For each MSFF module, the upper layer encoder output features (shallow features) ${{\mathbf{F}}_S} \in {\mathbb{R} ^{4n \times \left( {c/4} \right)}}$, the local layer encoder output features (local features) ${{\mathbf{F}}_L} \in {\mathbb{R} ^{n \times c}}$, and the lower layer encoder output features (deep features) ${{\mathbf{F}}_D} \in {\mathbb{R}^{\left( {n/4} \right) \times 4c}}$ are extracted, where $n$ and $c$ are the size of the receptive field and the dimension of the ${{\mathbf{F}}_L}$. To facilitate the aggregation of detailed features of features at different scales, ${{\mathbf{F}}_S}$ is downsampled to ${{\mathbf{F'}}_S} \in {\mathbb{R}^{n \times \left( {c/4} \right)}}$ and ${{\mathbf{F}}_D}$ is upsampled to ${{\mathbf{F'}}_D} \in {\mathbb{R}^{n \times 4c}}$ in order to match the size of ${{\mathbf{F}}_L}$.

Our feature fusion process proceeds as follows. We form shallow fusion weights ${\mathbf{e}_S}$ by merging ${{\mathbf{F'}}_S}$ and ${{\mathbf{F}}_L}$ as follows:
\begin{equation}
{\mathbf{e}_S} = \sigma \left( {{\mathcal{G}}\left( {{{{\mathbf{F'}}}_S}} \right) + {\mathcal{G}}\left( {{{\mathbf{F}}_L}} \right)} \right)
\end{equation}
where ${\mathcal{G}}\left(  \cdot  \right)$ denotes the global average pooling operation, and $\sigma $ is the \emph{Sigmoid} function.

The shallow fusion features ${{\mathbf{\tilde F}}_S}$ are then obtain. The shallow fusion weights modulate the shallow features ${\mathbf{e}_S}$ to produce the shallow \emph{fusion} features:
\begin{equation}
{{\mathbf{\tilde F}}_S} = {{\mathbf{F'}}_S} \otimes {\mathbf{e}_S}
\end{equation}
where $ \otimes $ is the element-wise multiplication operation.

Similarly, ${{\mathbf{F'}}_D}$ and ${{\mathbf{F}}_L}$ are merged to produce deep fusion weights:
\begin{equation}
{\mathbf{e}_D} = \sigma \left( {{\mathcal{G}}\left( {{{{\mathbf{F'}}}_D}} \right) \otimes {\mathcal{G}}\left( {{{\mathbf{F}}_L}} \right)} \right)
\end{equation}

And these modulate the deep features ${{\mathbf{F'}}_D}$ to yield the deep fusion features ${{\mathbf{\tilde F}}_D}$:
\begin{equation}
{{\mathbf{\tilde F}}_D} = {{\mathbf{F'}}_D} \otimes {\mathbf{e}_D}
\end{equation}

The resulting shallow fusion, deep fusion and local features are concatenated into one feature, which is passed through an \emph{MLP} layer to produce the output features ${{\mathbf{F}}_O} \in {\mathbb{R}^{n \times c}}$:
\begin{equation}
{{\mathbf{F}}_O} = {\mathcal{M}}\left( {{{{\mathbf{\tilde F}}}_S} \oplus {{\mathbf{F}}_L} \oplus {{{\mathbf{\tilde F}}}_D}} \right)
\end{equation}
where ${\mathcal{M}}\left( \cdot \right)$ is the \emph{MLP} operation, and $ \oplus $ is the concatenation operation.

\subsection{Adaptive hybrid loss}
\begin{figure}[!t]
    \centering
    \includegraphics[width=0.68\linewidth]{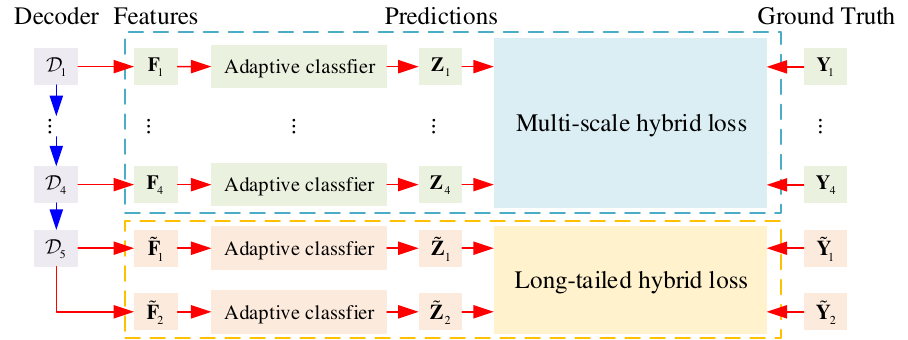}
    \caption{Architecture of the adaptive hybrid loss. Proposed loss includes two parts: multi-scale loss and long-tailed loss}
    \label{fig4}
\end{figure}

To deal effectively with long-tailed distribution and multi-scale feature of land-covers in remote sensing datasets, we construct an adaptive hybrid loss (AHL) module consisting of multi-scale loss and long-tailed loss. The key feature of the proposed AHL module is to simultaneously apply multi-classification heads to learn different classes of land-cover separately, and utilize adaptive weights to enhance the learning effect among different classes. The architecture of the adaptive hybrid loss can be seen in Figure~\ref{fig4}.

\subsubsection{Multi-scale loss}
In networks with the encoder-decoder structure, the resolution of the receptive field gradually decreases as the number of network layers deepens, while the feature dimension of the remaining points gradually increases. Land-cover scales in remote sensing datasets vary greatly, and there may be a 50-fold difference between the scales of the largest land-covers (e.g., buildings) and the smallest land-covers (e.g., street lights) in some datasets. Therefore, small-scale land-covers may be lost at lower resolutions, while large-scale land-covers are easily split into different parts and learned features separately at higher resolutions, leading to misclassification.

In summary, the classification accuracy of features at all scales is taken into account, we hope that both low- and high-resolution features output by the decoder can be effectively classified. Therefore, the multi-scale loss module is proposed to enhance the learning ability of the network for the various-scale land-covers.

Let ${\mathbf{F}}^{i}$ denote the feature output of the $i$th decoder layer. We have that ${{\mathbf{F}}^{i}} \in {\mathbb{R}^{{4^{i - 5}}N \times {4^{5 - i}}C}}$. Thus the feature output of the fifth, i.e., the last decoder layer is ${{\mathbf{F}}^{5}} \in {\mathbb{R}^{N \times C}}$, where $N$ and $C$ are the size of the receptive field and the dimension of the features.

For the low-resolution features output from the first four layers, a simple classifier consisting of three fully connected layers is used to obtain the prediction map ${{\mathbf{Z}}^{i}} = \left[ {{\mathbf{z}}_1^{i},{\mathbf{z}}_2^{i}, \ldots ,{\mathbf{z}}_N^{i}} \right] \in {\mathbb{R}^{{4^{i - 5}}N \times L}}$ at the corresponding resolutions, where $L$ is the number of classes. Since the classification of different land-covers at different scales is not the same, an adaptive weight ${\omega ^{i}} \in {\mathbb{R}^{1 \times L}}$ is applied to each classifier to automatically extract the features with the optimal classification accuracy at the current resolution:
\begin{equation}
{{\mathbf{Z}}^{i}} = {\omega ^{i}} \otimes {\mathcal{C}}\left( {{{\mathbf{F}}^{i}}} \right)
\end{equation}
where ${\mathcal{C}}\left( \cdot \right)$ is the classifier.

Then the resulting category probability map is used with the low-resolution truth map ${{\mathbf{Y}}^{i}} = \left[ {{\mathbf{y}}_1^{i},{\mathbf{y}}_2^{i}, \ldots ,{\mathbf{y}}_N^{i}} \right] \in {\mathbb{R}^{{4^{i - 5}}N \times L}}_{}^{}$ generated by down-sampling the truth maps to the corresponding resolutions to compute the loss utilizing the cross-entropy function:
\begin{equation}
{{\mathcal{L}}^{i}} = - \sum\limits_{j = 1}^N {{\mathbf{y}}_j^{i}{{\log }_2}\left( {{\mathbf{z}}_j^{i}} \right)}
\end{equation}

The final multi-scale hybrid loss is the sum of losses at each resolution, scaled to the same scale according to adaptive weights:
\begin{equation}
{{\mathcal{L}}^{scale}} = \sum\limits_{i = 1}^4 {\frac{{{{\mathcal{L}}^{i}}}}{{{4^{i - 1}}\left\| {{\omega ^{i}}} \right\|}}}
\end{equation}
where $\left\| \cdot \right\|$ is the ${l_2}$ norm.

\subsubsection{Long-tailed loss}
In addition to the non-uniform distribution on the object scale described in the previous section, another type of non-uniform distribution often exists in remote sensing scenarios: the non-uniform distribution on the number of land-covers, i.e., the long-tailed distribution.

Long-tailed distributions can lead to differences in learning efficiency between different classes of land-covers, making it difficult for tail classes to learn effective features and resulting in overfitting phenomena. Long-tailed distributions can lead to differences in learning efficiency between different classes of land-covers, making it difficult for tail classes to learn effective features and resulting in overfitting phenomena. To solve the above phenomenon, the long-tailed loss is proposed to enhance the learning ability of the network for the small classes.

The final feature ${{\mathbf{F}}^{5}}$ is fed into two classifiers constructed from fully connected layers to obtain prediction maps ${{\mathbf{\tilde Z}}^{i}} = \left[ {{\mathbf{\tilde z}}_1^{i},{\mathbf{\tilde z}}_2^{i}, \ldots ,{\mathbf{\tilde z}}_N^{i}} \right] \in {\mathbb{R}^{N \times L}}$, $i = 1,2$. Similar to multi-scale loss, an adaptive weight ${\tilde \omega ^{i}} \in {\mathbb{R}^{1 \times L}}$ is applied to each classifier:
\begin{equation}
{{\mathbf{\tilde Z}}^{i}} = {\tilde \omega ^{i}} \otimes {\mathcal{C}}\left( {{{\mathbf{F}}^{5}}} \right)
\end{equation}

The final classification result is the weighted sum of all the classifiers:
\begin{equation}
{{\mathbf{Z}}^{tail}} = \sum\limits_{i = 1}^2 {{{{\mathbf{\tilde Z}}}^{i}}}
\end{equation}

For the loss function, we want the first classifier to learn all classes of land-covers, while the second classifier learns only the categorical features of the tail classes. Therefore how to make the two classifiers cooperate is the focus of the loss function design.

Compared with the common cross-entropy loss function, the root mean squared loss function can better suppress the output probability of non-target classes and reduce the interaction of results between different classifiers:
\begin{equation}
{{\tilde{\mathcal{L}}}^{i}} = \frac{1}{N}\sum\limits_{j = 1}^N {{{\left( {{\mathbf{\tilde y}}_j^{i} - {\mathbf{\tilde z}}_j^{i}} \right)}^2}}
\end{equation}
where: ${{\mathbf{\tilde Y}}^{1}} = \left[ {{\mathbf{\tilde y}}_1^{1},{\mathbf{\tilde y}}_2^{1}, \ldots ,{\mathbf{\tilde y}}_N^{1}} \right] \in {\mathbb{R}^{N \times L}}$ represents the ground truth of all land-covers and ${{\mathbf{\tilde Y}}^{2}} = \left[ {{\mathbf{\tilde y}}_1^{2},{\mathbf{\tilde y}}_2^{2}, \ldots ,{\mathbf{\tilde y}}_N^{2}} \right] \in {\mathbb{R}^{N \times L}}$ represents the ground truth of tail land-covers (with the head land-covers set to $\mathbf{0}$). In this study, the classes that account for less than 5\% of the points in the training set are used as the tail classes.

The final long-tailed loss is the sum of losses at each classifiers, scaled to the same scale according to adaptive weights:
\begin{equation}
{{\mathcal{L}}^{tail}} = \sum\limits_{i = 1}^2 {\frac{{{{{\tilde{\mathcal{L}}}}^{i}}}}{{\left\| {{{\tilde \omega }^{ith}}} \right\|}}}
\end{equation}

\subsubsection{Adaptive hybrid loss}
In order to balance the learning ability of the network in response to multi-scale and long-tailed distribution land covers, the final loss of the network should include both multi-scale loss and long-tailed loss. Since the long-tailed loss directly determines the network outcomes, and the multi-scale loss mainly affects the interpretability of the output features of each decoder layer, a weight ${\lambda}$ is added to the multi-scale loss to deflate its learning rate, and the final adaptive hybrid loss is equal to the sum of the two loss functions.
\begin{equation}
{\mathcal{L}} = \lambda{{\mathcal{L}}^{scale}} + {{\mathcal{L}}^{tail}}
\end{equation}

\section{Experimental results and analysis}
\subsection{Data description and hyperparameter setting}
\subsubsection{Datasets}
\begin{figure}[!t]
    \centering
    \subfloat[]
    {
        \includegraphics[width=0.30\linewidth]{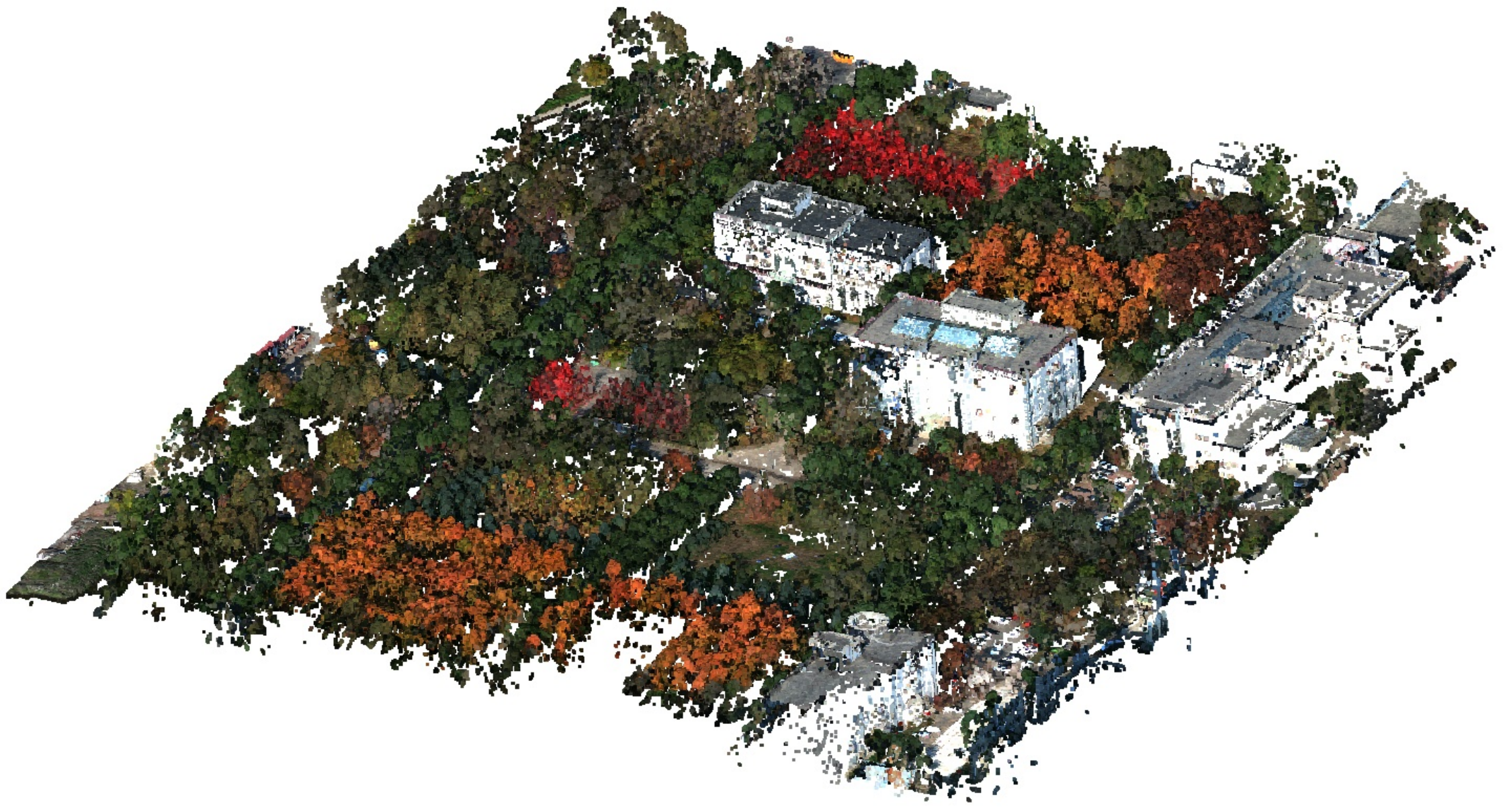}
    }
    \hspace{0.02\textwidth}
    \subfloat[]{
        \includegraphics[width=0.30\linewidth]{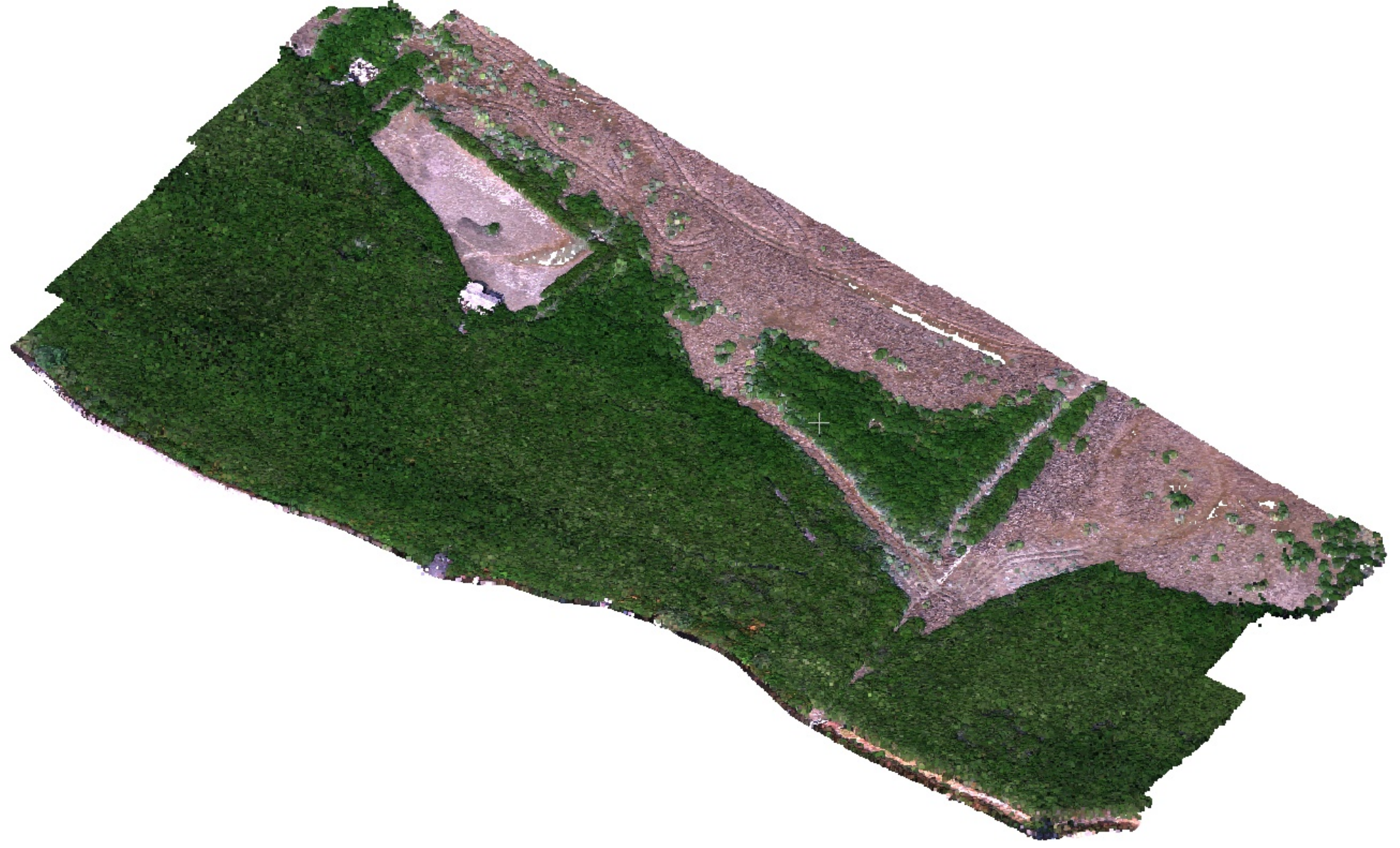}
    }
    \hspace{0.02\textwidth}
    \subfloat[]
    {
        \includegraphics[width=0.30\linewidth]{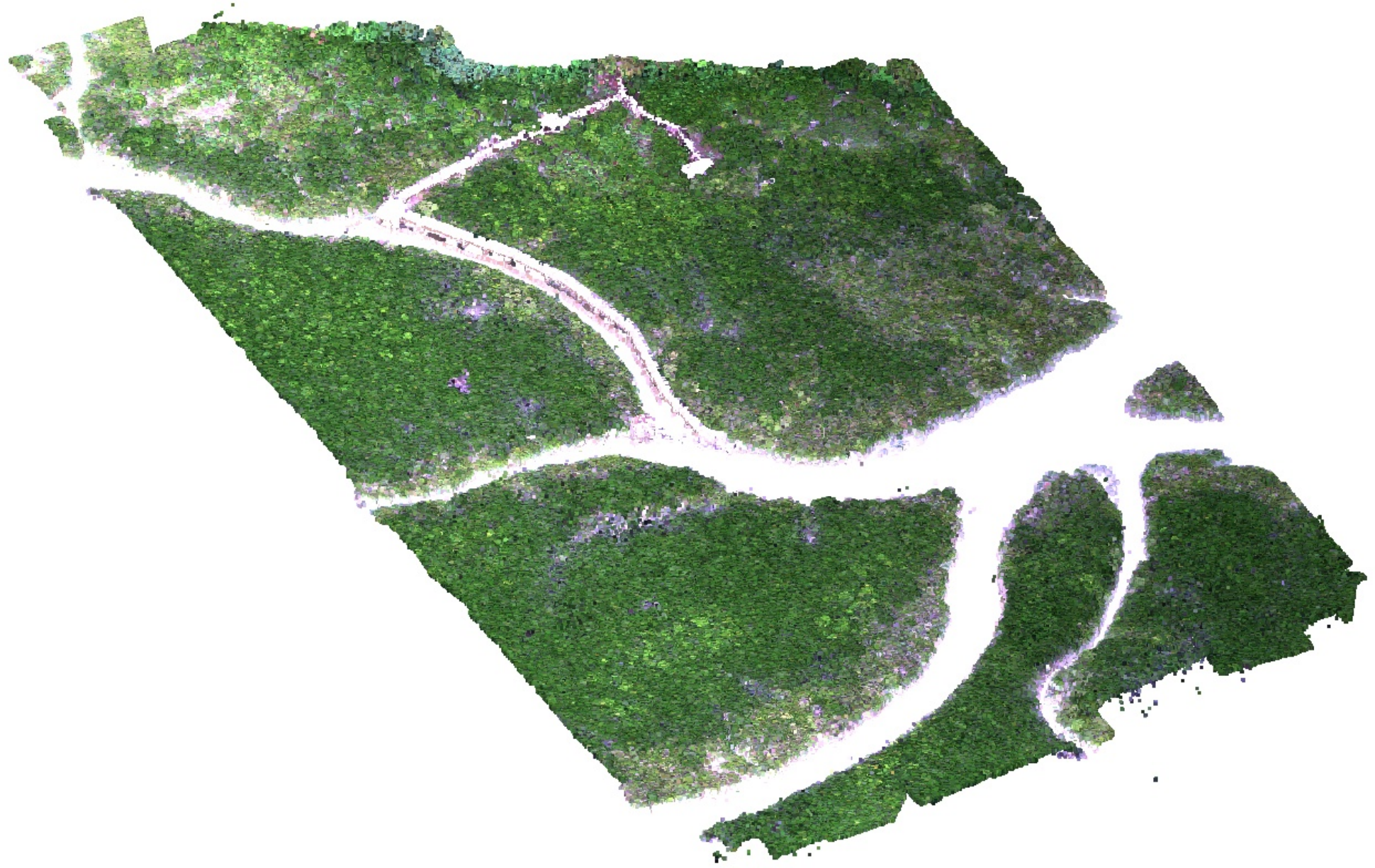}
    }
    \caption{Multispectral point cloud datasets used in the experiment. (a) HIT dataset; (b) ZJK dataset; (c) SK dataset.}
    \label{fig5}
\end{figure}

In this section, the effectiveness of the proposed module is evaluated on three real-life MPC datasets: Harbin Institute of Technology (HIT) dataset, Zhangjiangkou (ZJK) dataset, and Shankou (SK) dataset. These datasets are obtained using the 3D reconstruction method proposed by Wang et al.~\cite{48,52}, as shown in Figure~\ref{fig5}.

\begin{table}[!t]
    \centering
    \scriptsize
    \caption{Comparison of the classification accuracies among the proposed method and the baselines using the HIT datasets.}
    \label{tab1}
    \tabcolsep 2pt
    \begin{tabular}{lccccccccc}
    \toprule
        Model & \makecell{RandLA-Net \\ (2020)} & \makecell{BAAF-Net \\ (2021)} & \makecell{CGA-Net \\ (2021)} & \makecell{SCF-Net \\ (2021)} & \makecell{RFFS-Net \\ (2022)} & \makecell{GeoSegNet \\ (2023)} & \makecell{EyeNet \\ (2023)} & \makecell{Hu et al. \\ (2023)} & Ours\\\hline
        LT & 85.39$\pm$1.22 & 95.01$\pm$0.31 & \underline{96.09$\pm$2.35} & \textbf{97.60$\pm$1.23} & 79.89$\pm$0.79 & 54.35$\pm$24.38 & 26.52$\pm$8.26 & 85.24$\pm$3.17 & 79.46$\pm$2.25 \\
        BD & 86.46$\pm$14.41 & \underline{95.28$\pm$0.74} & 65.01$\pm$45.88 & \textbf{96.69$\pm$0.23} & 84.26$\pm$3.06 & 87.23$\pm$0.87 & 0.06$\pm$0.02 & 89.53$\pm$3.62 & 92.63$\pm$2.21 \\
        PP & 87.39$\pm$4.63 & 77.35$\pm$1.26 & \underline{91.57$\pm$3.03} & 86.41$\pm$2.36 & 76.21$\pm$5.89 & 48.35$\pm$36.49 & 0.64$\pm$0.09 & 75.57$\pm$10.89 & \textbf{92.28$\pm$2.83} \\
        GR & 87.97$\pm$5.55 & \underline{94.24$\pm$0.71} & 66.07$\pm$43.12 & \textbf{97.01$\pm$0.32} & 81.94$\pm$6.03 & 1.12$\pm$1.58 & 2.45$\pm$0.68 & 85.53$\pm$1.82 & 93.70$\pm$1.56 \\
        AP & 66.95$\pm$10.87 & 80.48$\pm$0.13 & \textbf{90.66$\pm$1.85} & \underline{87.37$\pm$2.03} & 69.42$\pm$4.06 & 0.88$\pm$1.22 & 0.00$\pm$0.00 & 50.24$\pm$33.00 & 64.72$\pm$7.63 \\
        PN & 52.75$\pm$3.64 & 0.00$\pm$0.00 & 17.84$\pm$25.14 & 7.23$\pm$10.01 & 48.92$\pm$4.67 & 0.00$\pm$0.00 & 22.26$\pm$7.75 & \underline{74.76$\pm$7.99} & \textbf{93.26$\pm$4.59} \\
        GT & \textbf{96.43$\pm$1.65} & 80.77$\pm$12.10 & 63.69$\pm$45.04 & 95.01$\pm$0.42 & 73.46$\pm$13.34 & 0.00$\pm$0.00 & 0.00$\pm$0.00 & 83.84$\pm$10.18 & \underline{97.37$\pm$0.83} \\
        BP & 41.77$\pm$9.26 & 0.00$\pm$0.00 & 1.69$\pm$2.40 & 12.37$\pm$6.38 & 11.91$\pm$7.56 & 0.00$\pm$0.00 & 0.00$\pm$0.00 & \underline{63.26$\pm$31.03} & \textbf{87.56$\pm$1.81} \\
        CA & 80.59$\pm$7.52 & 1.35$\pm$1.45 & \underline{88.88$\pm$0.71} & 88.67$\pm$1.66 & 47.15$\pm$5.13 & 0.00$\pm$0.00 & 0.00$\pm$0.00 & 74.52$\pm$17.22 & \textbf{96.23$\pm$1.02} \\
        SH & 37.27$\pm$5.60 & 0.00$\pm$0.00 & 0.94$\pm$1.00 & 0.59$\pm$0.84 & 0.10$\pm$0.14 & 0.00$\pm$0.00 & 0.00$\pm$0.00 & \underline{40.49$\pm$19.48} & \textbf{90.29$\pm$3.04} \\
        FT & \underline{84.09$\pm$16.08} & 0.00$\pm$0.00 & 55.58$\pm$39.42 & 81.06$\pm$9.31 & 20.27$\pm$28.67 & 0.00$\pm$0.00 & 0.00$\pm$0.00 & 71.74$\pm$10.38 & \textbf{99.70$\pm$0.15} \\
        SL & 29.54$\pm$2.86 & 0.00$\pm$0.00 & 0.00$\pm$0.00 & 0.04$\pm$0.06 & 0.00$\pm$0.00 & 0.00$\pm$0.00 & 0.00$\pm$0.00 & \underline{38.81$\pm$21.09} & \textbf{95.16$\pm$1.47} \\
        OA & 81.04$\pm$2.71 & 77.98$\pm$0.49 & \underline{86.30$\pm$0.87} & 85.22$\pm$0.09 & 72.80$\pm$0.58 & 39.77$\pm$8.25 & 7.82$\pm$2.28 & 78.84$\pm$3.70 & \textbf{87.07$\pm$1.11} \\
        AA & \underline{69.72$\pm$1.45} & 43.71$\pm$1.07 & 59.05$\pm$5.12 & 62.51$\pm$1.12 & 49.46$\pm$3.03 & 15.99$\pm$3.43 & 4.33$\pm$1.26 & 69.46$\pm$5.49 & \textbf{90.20$\pm$1.05} \\
        kappa & 77.55$\pm$3.08 & 73.13$\pm$0.64 & \underline{83.46$\pm$1.11} & 82.15$\pm$0.11 & 68.03$\pm$0.53 & 24.18$\pm$10.63 & -12.97$\pm$2.98 & 75.14$\pm$4.33 & \textbf{84.89$\pm$1.28} \\
        mIoU & \underline{55.82$\pm$2.22} & 35.63$\pm$0.99 & 51.74$\pm$5.02 & 54.71$\pm$1.17 & 38.42$\pm$1.35 & 8.59$\pm$2.59 & 1.92$\pm$0.52 & 54.38$\pm$5.17 & \textbf{67.66$\pm$3.07} \\
        head avg. & 82.83$\pm$2.31 & 88.47$\pm$0.11 & \textbf{95.76$\pm$1.64} & \underline{93.02$\pm$0.69} & 78.34$\pm$1.47 & 38.39$\pm$8.22 & 5.94$\pm$1.73 & 77.22$\pm$7.99 & 84.56$\pm$1.61 \\
        tail avg. & 60.35$\pm$0.88 & 11.73$\pm$1.88 & 57.09$\pm$29.88 & 40.71$\pm$2.41 & 28.83$\pm$6.23 & 0.00$\pm$0.00 & 3.18$\pm$1.11 & \underline{63.92$\pm$5.70} & \textbf{94.22$\pm$1.13} \\
        head min. & 61.96$\pm$5.10 & 77.35$\pm$1.26 & \textbf{90.10$\pm$1.40} & \underline{86.41$\pm$2.36} & 69.42$\pm$4.06 & 0.00$\pm$0.00 & 0.00$\pm$0.00 & 45.18$\pm$26.03 & 64.72$\pm$7.63 \\
        tail min. & 28.36$\pm$1.91 & 0.00$\pm$0.00 & \underline{32.15$\pm$45.47} & 0.00$\pm$0.00 & 0.00$\pm$0.00 & 0.00$\pm$0.00 & 0.00$\pm$0.00 & 24.62$\pm$13.35 & \textbf{87.06$\pm$1.73} \\
    \bottomrule
    \end{tabular}
    \end{table}

\begin{table}[!t]
    \centering
    \scriptsize
    \caption{Comparison of the classification accuracies among the proposed method and the baselines using the ZJK datasets.}
    \label{tab2}
    \tabcolsep 2pt
    \begin{tabular}{lccccccccc}
    \toprule
        Model & \makecell{RandLA-Net \\ (2020)} & \makecell{BAAF-Net \\ (2021)} & \makecell{CGA-Net \\ (2021)} & \makecell{SCF-Net \\ (2021)} & \makecell{RFFS-Net \\ (2022)} & \makecell{GeoSegNet \\ (2023)} & \makecell{EyeNet \\ (2023)} & \makecell{Hu et al. \\ (2023)} & Ours\\\hline
        TF & \textbf{99.87$\pm$0.11} & 97.63$\pm$0.30 & 97.39$\pm$0.40 & 97.55$\pm$0.92 & 56.47$\pm$0.83 & 82.45$\pm$23.02 & 95.04$\pm$2.40 & 94.78$\pm$6.73 & \underline{99.25$\pm$0.48} \\
        BD & 53.45$\pm$5.26 & 98.03$\pm$0.50 & \underline{98.55$\pm$0.20} & 98.16$\pm$1.47 & 5.56$\pm$7.74 & 0.00$\pm$0.00 & 0.00$\pm$0.00 & 1.59$\pm$2.04 & \textbf{99.18$\pm$0.78} \\
        BG & \underline{88.94$\pm$5.94} & 0.00$\pm$0.00 & 0.00$\pm$0.00 & 0.00$\pm$0.00 & 54.16$\pm$6.15 & 37.02$\pm$26.71 & 5.25$\pm$1.46 & 85.92$\pm$11.32 & \textbf{95.66$\pm$3.63} \\
        AC & \underline{7.54$\pm$6.65} & 0.00$\pm$0.00 & 0.00$\pm$0.00 & 0.00$\pm$0.00 & 0.00$\pm$0.00 & 0.00$\pm$0.00 & 0.43$\pm$0.17 & 5.45$\pm$5.73 & \textbf{12.77$\pm$8.49} \\
        TP & 34.59$\pm$20.57 & 49.21$\pm$5.23 & 50.85$\pm$2.83 & 51.70$\pm$1.57 & 3.50$\pm$4.95 & 7.51$\pm$10.61 & 0.45$\pm$0.25 & \underline{54.97$\pm$38.90} & \textbf{98.61$\pm$0.69} \\
        AM & \underline{51.68$\pm$3.82} & 0.00$\pm$0.00 & 0.00$\pm$0.00 & 0.00$\pm$0.00 & 3.83$\pm$2.98 & 0.00$\pm$0.00 & 5.35$\pm$2.55 & 23.35$\pm$22.83 & \textbf{77.15$\pm$8.96} \\
        KO & 15.16$\pm$11.78 & 0.00$\pm$0.00 & 0.00$\pm$0.00 & 0.00$\pm$0.00 & 10.22$\pm$14.46 & 1.29$\pm$1.83 & 0.00$\pm$0.00 & \underline{38.73$\pm$18.31} & \textbf{70.09$\pm$8.95} \\
        OA & \underline{79.77$\pm$1.27} & 74.02$\pm$0.14 & 74.02$\pm$0.43 & 74.13$\pm$0.60 & 40.48$\pm$1.55 & 54.75$\pm$16.53 & 60.30$\pm$1.31 & 71.61$\pm$4.63 & \textbf{91.73$\pm$0.28} \\
        AA & \underline{50.18$\pm$3.75} & 34.98$\pm$0.69 & 35.26$\pm$0.48 & 35.34$\pm$0.29 & 19.11$\pm$3.98 & 18.32$\pm$7.82 & 15.22$\pm$0.19 & 43.54$\pm$7.07 & \textbf{78.96$\pm$1.98} \\
        kappa & \underline{61.59$\pm$3.07} & 57.01$\pm$0.28 & 57.08$\pm$0.59 & 57.20$\pm$0.73 & 16.77$\pm$2.81 & 7.91$\pm$20.04 & 26.45$\pm$1.30 & 49.24$\pm$8.85 & \textbf{85.82$\pm$0.54} \\
        mIoU & \underline{39.35$\pm$3.33} & 34.88$\pm$0.70 & 35.20$\pm$0.50 & 35.18$\pm$0.36 & 14.61$\pm$3.40 & 12.08$\pm$5.65 & 12.67$\pm$0.03 & 27.52$\pm$5.19 & \textbf{66.93$\pm$1.13} \\
        head avg. & \underline{62.45$\pm$1.14} & 48.91$\pm$0.13 & 48.99$\pm$0.15 & 48.93$\pm$0.32 & 29.05$\pm$3.35 & 29.87$\pm$12.38 & 25.18$\pm$0.26 & 46.93$\pm$3.14 & \textbf{76.72$\pm$1.74} \\
        tail avg. & 33.81$\pm$7.28 & 16.40$\pm$1.74 & 16.95$\pm$0.94 & 17.23$\pm$0.52 & 5.85$\pm$4.88 & 2.93$\pm$3.28 & 1.93$\pm$0.78 & \underline{39.02$\pm$12.76} & \textbf{81.95$\pm$4.59} \\
        head min. & \underline{7.54$\pm$6.65} & 0.00$\pm$0.00 & 0.00$\pm$0.00 & 0.00$\pm$0.00 & 0.00$\pm$0.00 & 0.00$\pm$0.00 & 0.00$\pm$0.00 & 1.09$\pm$1.34 & \textbf{12.77$\pm$8.49} \\
        tail min. & 11.50$\pm$12.40 & 0.00$\pm$0.00 & 0.00$\pm$0.00 & 0.00$\pm$0.00 & 0.00$\pm$0.00 & 0.00$\pm$0.00 & 0.00$\pm$0.00 & \underline{12.29$\pm$10.16} & \textbf{68.26$\pm$9.32} \\
    \bottomrule
    \end{tabular}
    \end{table}

\begin{table}[!t]
    \centering
    \scriptsize
    \caption{Comparison of the classification accuracies among the proposed method and the baselines using the SK datasets.}
    \label{tab3}
    \tabcolsep 2pt
    \begin{tabular}{lccccccccc}
    \toprule
        Model & \makecell{RandLA-Net \\ (2020)} & \makecell{BAAF-Net \\ (2021)} & \makecell{CGA-Net \\ (2021)} & \makecell{SCF-Net \\ (2021)} & \makecell{RFFS-Net \\ (2022)} & \makecell{GeoSegNet \\ (2023)} & \makecell{EyeNet \\ (2023)} & \makecell{Hu et al. \\ (2023)} & Ours\\\hline
        BR & \textbf{96.19$\pm$3.15} & 80.69$\pm$0.63 & 80.87$\pm$0.76 & 80.37$\pm$3.22 & 36.53$\pm$8.65 & 49.55$\pm$36.03 & 6.54$\pm$7.90 & 52.77$\pm$38.16 & \underline{94.55$\pm$3.88} \\
        BG & 63.97$\pm$25.78 & 21.45$\pm$15.13 & 34.36$\pm$0.66 & 29.89$\pm$6.73 & 31.18$\pm$12.83 & 31.18$\pm$19.31 & \textbf{94.29$\pm$3.37} & 48.95$\pm$35.96 & \underline{83.82$\pm$2.15} \\
        RS & 31.68$\pm$5.24 & 0.62$\pm$0.88 & 4.81$\pm$0.92 & 0.00$\pm$0.00 & 14.17$\pm$19.52 & 8.29$\pm$6.58 & \textbf{89.01$\pm$4.47} & 31.93$\pm$23.02 & \underline{78.12$\pm$10.81} \\
        AM & 35.03$\pm$24.78 & 0.57$\pm$0.71 & 0.52$\pm$0.24 & 0.99$\pm$1.22 & 20.39$\pm$25.46 & 0.35$\pm$0.50 & \underline{57.63$\pm$6.44} & 23.98$\pm$33.02 & \textbf{77.49$\pm$3.77} \\
        KO & 21.15$\pm$6.37 & 0.00$\pm$0.00 & 0.00$\pm$0.00 & 0.00$\pm$0.00 & 2.01$\pm$2.85 & 0.95$\pm$1.34 & 6.86$\pm$9.65 & \underline{41.09$\pm$17.99} & \textbf{60.03$\pm$22.42} \\
        BE & 80.72$\pm$6.84 & 39.73$\pm$2.12 & 40.76$\pm$2.08 & 45.99$\pm$4.35 & 20.50$\pm$15.35 & 35.20$\pm$44.91 & \textbf{97.83$\pm$0.70} & 76.12$\pm$32.01 & \underline{97.21$\pm$1.32} \\
        AC & 1.60$\pm$1.55 & 0.00$\pm$0.00 & 0.00$\pm$0.00 & 0.00$\pm$0.00 & 0.00$\pm$0.01 & 0.30$\pm$0.29 & 9.99$\pm$13.92 & \underline{14.98$\pm$18.00} & \textbf{77.53$\pm$17.11} \\
        BT & 0.00$\pm$0.00 & 90.71$\pm$0.04 & \underline{93.98$\pm$1.32} & 88.63$\pm$4.03 & 0.00$\pm$0.00 & 0.00$\pm$0.00 & \textbf{100.00$\pm$0.00} & 0.00$\pm$0.00 & \textbf{100.00$\pm$0.00} \\
        OA & \underline{68.89$\pm$5.22} & 44.19$\pm$3.81 & 47.98$\pm$0.55 & 46.19$\pm$0.67 & 27.95$\pm$3.58 & 32.19$\pm$11.32 & 45.75$\pm$4.08 & 45.99$\pm$8.43 & \textbf{86.91$\pm$2.49} \\
        AA & 41.29$\pm$1.73 & 29.22$\pm$2.19 & 31.91$\pm$0.39 & 30.73$\pm$0.62 & 15.60$\pm$2.18 & 15.73$\pm$2.92 & \underline{50.16$\pm$12.20} & 36.23$\pm$3.25 & \textbf{83.59$\pm$4.65} \\
        kappa & \underline{56.39$\pm$5.85} & 32.08$\pm$3.86 & 36.00$\pm$0.52 & 34.13$\pm$0.71 & 18.31$\pm$3.50 & 5.82$\pm$1.78 & 44.35$\pm$8.98 & 33.16$\pm$7.26 & \textbf{82.03$\pm$3.30} \\
        mIoU & 29.84$\pm$2.56 & 29.16$\pm$2.26 & \underline{31.90$\pm$0.40} & 28.03$\pm$3.32 & 11.88$\pm$2.21 & 7.08$\pm$0.65 & 26.46$\pm$3.36 & 18.66$\pm$2.14 & \textbf{67.80$\pm$4.46} \\
        head avg. & 56.72$\pm$2.56 & 25.83$\pm$4.01 & 30.14$\pm$0.44 & 27.81$\pm$0.70 & 25.57$\pm$4.21 & 22.34$\pm$5.53 & \underline{61.87$\pm$3.82} & 39.41$\pm$7.23 & \textbf{83.49$\pm$0.25} \\
        tail avg. & 25.87$\pm$2.26 & 32.61$\pm$0.54 & 33.69$\pm$0.71 & 33.66$\pm$1.34 & 5.63$\pm$4.03 & 9.11$\pm$11.03 & \underline{53.67$\pm$3.29} & 33.05$\pm$13.02 & \textbf{83.69$\pm$9.55} \\
        head min. & \underline{19.93$\pm$3.43} & 0.05$\pm$0.07 & 0.52$\pm$0.24 & 0.00$\pm$0.00 & 0.24$\pm$0.35 & 0.35$\pm$0.50 & 6.54$\pm$7.90 & 0.69$\pm$0.52 & \textbf{71.04$\pm$0.79} \\
        tail min. & 0.00$\pm$0.00 & 0.00$\pm$0.00 & 0.00$\pm$0.00 & 0.00$\pm$0.00 & 0.00$\pm$0.00 & 0.00$\pm$0.00 & \underline{0.11$\pm$0.12} & 0.00$\pm$0.00 & \textbf{60.03$\pm$22.42} \\
        \bottomrule
    \end{tabular}
    \end{table}

The HIT dataset is acquired from the campus of Harbin Institute of Technology. The scene primarily consists of diverse vegetation, with a few buildings and vehicles present. We manually categorized the scene into 12 categories: leafy trees (LT), buildings (BD), poplars (PP), grass (GR), asphalt pavement (AP), pines (PN), gray leafy trees (GT), brick pavement (BP), cars (CA), shrubs (SH), flowering trees (FT), and street lamps (SL).

The ZJK dataset is acquired near the Zhangjiangkou Mangrove Nature Research Center, Fujian Province. This scene contains mainly a variety of vegetation, with large areas of tidal flats. We manually categorized the scene into 7 categories: tidal flats (TF), buildings (BD), B. gymnorhiza (BG), A. corniculatum (AC), timber pavement (TP), A. marina (AM), K. obovata (KO). Because the vegetation in this scene is too dense, the proportion of labeled data is low.

The SK dataset is acquired in the Shankou Mangrove Ecological Nature Reserve, Guangxi Zhuang Autonomous Region. This scene contains mainly a variety of vegetation. We manually categorized the scene into 8 categories: bridge (BR), B. gymnorhiza (BG), A. marina (AM), K. obovata (KO), bare earth (BE), A. corniculatum (AC), and boats (BT). Because the vegetation in this scene is too dense, the proportion of labeled data is low.

\subsubsection{Implementation details}
All experiments are run on a single RTX 3050 GPU and Intel i7-12700 CPU, the operating system is Windows 10. The proposed method is implemented with the PyTorch framework and other methods are implemented as described in the corresponding articles. The receptive field is set to 4096 and the batch size to 16. The maximum number of training epochs for all models is 100 until the model converges. The learning rate is set to 0.005 and at each epoch is reduced by 5\%. The decay rate is set to 0.0001.

\subsubsection{Evaluation metrics}
To qualitatively evaluate the classification performance of the proposed network, the overall accuracy (OA), average accuracy over classes (AA), kappa and mean intersection over union (mIoU) are used in this article.

Further, to evaluate the optimization effect of the proposed method for long-tailed distribution data, the average precision of the head categories (head avg.), the average precision of the tail categories (tail avg.), the minimum precision of the head categories (head min.), and the minimum precision of the tail categories (tail min.) are calculated, respectively.

\subsection{Evaluation results and analysis}
The proposed method is evaluated on three datasets, the HIT, the ZJK, and the SK datasets, and compared with the existing approaches. Eight representative methods such as RandLA-Net~\cite{34}, BAAF-Net~\cite{36}, CGA-Net~\cite{35}, SCF-Net~\cite{37}, RFFS-Net~\cite{49}, GeoSegNet~\cite{50}, EyeNet~\cite{38}, and Hu et al.~\cite{51} works in the classification of point clouds and are still widely used (open source), are used as solid baselines to evaluate our method.

Table~\ref{tab1}, Table~\ref{tab2} and Table~\ref{tab3} report class-specific accuracy, AA, OA, kappa, mIoU, and long-tailed distribution related metrics for all methods on the HIT, ZJK, and SK datasets. It can be seen that the proposed method achieves the best OA, AA, kappa and mIoU in all three datasets with significant improvement over the comparison methods. For example, it is observed in Table II that the proposed method achieves the highest OA and improves 11.96\% compared to the highest value in the comparison method, which appears in the RandLA-Net. Similar results are observed on the other two datasets.

As shown in Table~\ref{tab2} and Table~\ref{tab3}, due to the sparse manually labeled truth values in these two datasets, most of the MPC classification methods are difficult to learn the classification features effectively in the face of such truth values, which leads to the classification accuracy of 0\% in some of the disadvantaged categories, e.g., the classification accuracy of SCF-Net is 0\% for four categories in the ZJK dataset, namely B. gymnorhiza, A. corniculatum, C. sinensis and K. obovata in the ZJK dataset; while the proposed method achieves the highest classification accuracy in most of the categories in the three datasets, which verifies the effectiveness of the proposed method.

On the other hand, in the long-tail distribution metrics, the proposed method achieves the optimal tail category metrics (tail avg. and tail min.) in all three datasets, however, the proposed method fails to achieve the optimal head category metrics on the HIT dataset, with the head avg. falling below the optimal metrics by 11.20\% and the head mean. falling below the optimal metrics by 25.42\%. Observing the CGA-Net method that achieves the optimal head metrics on the HIT dataset, it can be seen that the method learns the head categories very well, but learns most tail categories very poorly, with a classification accuracy of less than 2\%, which ultimately leads to a low OA of the method; while the proposed method learns some head categories poorly, but achieves the optimal learning results for all tail categories, the Thus a higher OA is finally obtained and overall better classification results are obtained.

\begin{figure*}[!t]
    \centering
    \begin{minipage}[b]{0.30\textwidth}
        \centering
        \subfloat[]{
            \begin{minipage}[t]{0.95\textwidth}
                \includegraphics[width=\linewidth]{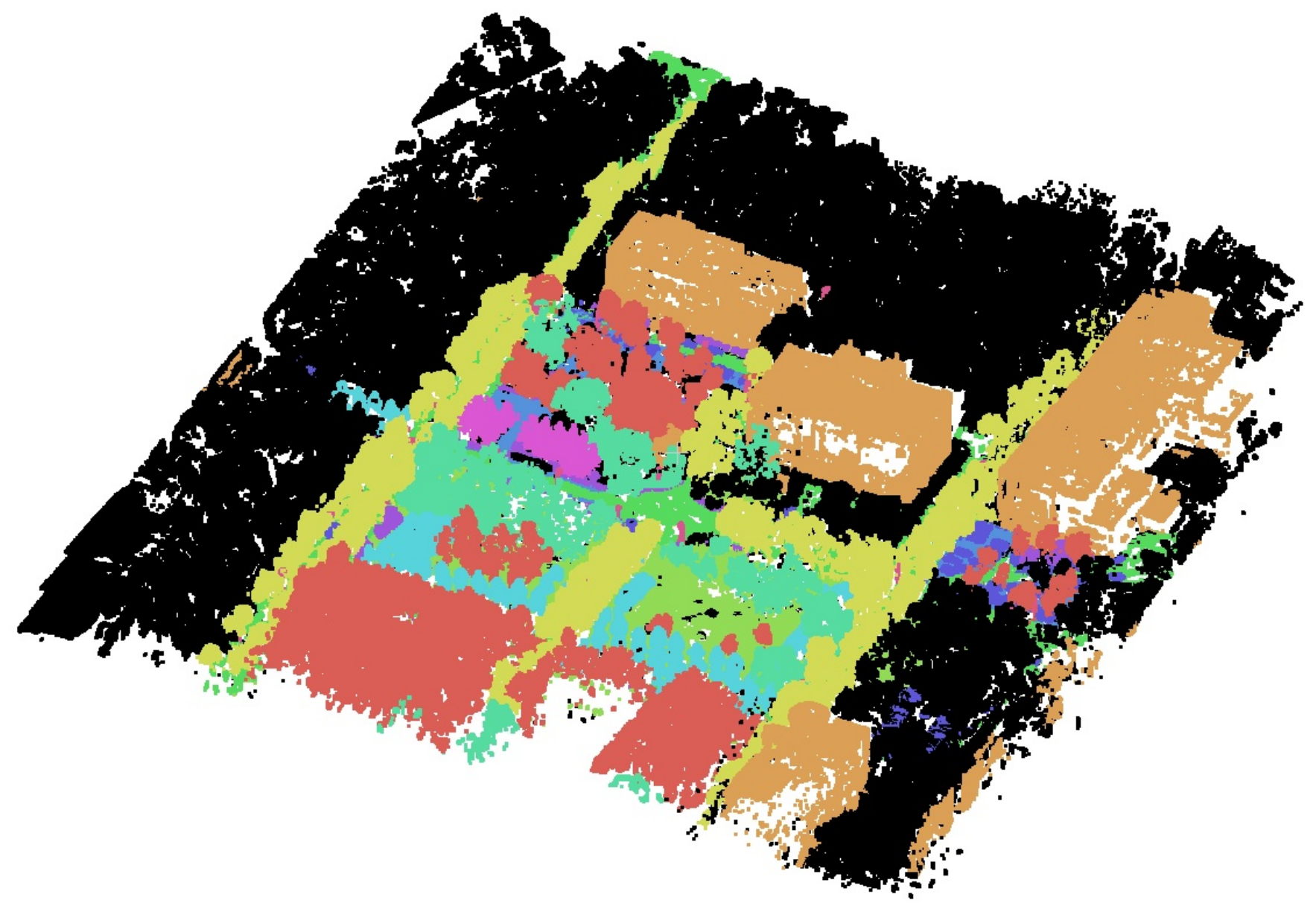}
                \includegraphics[width=\linewidth]{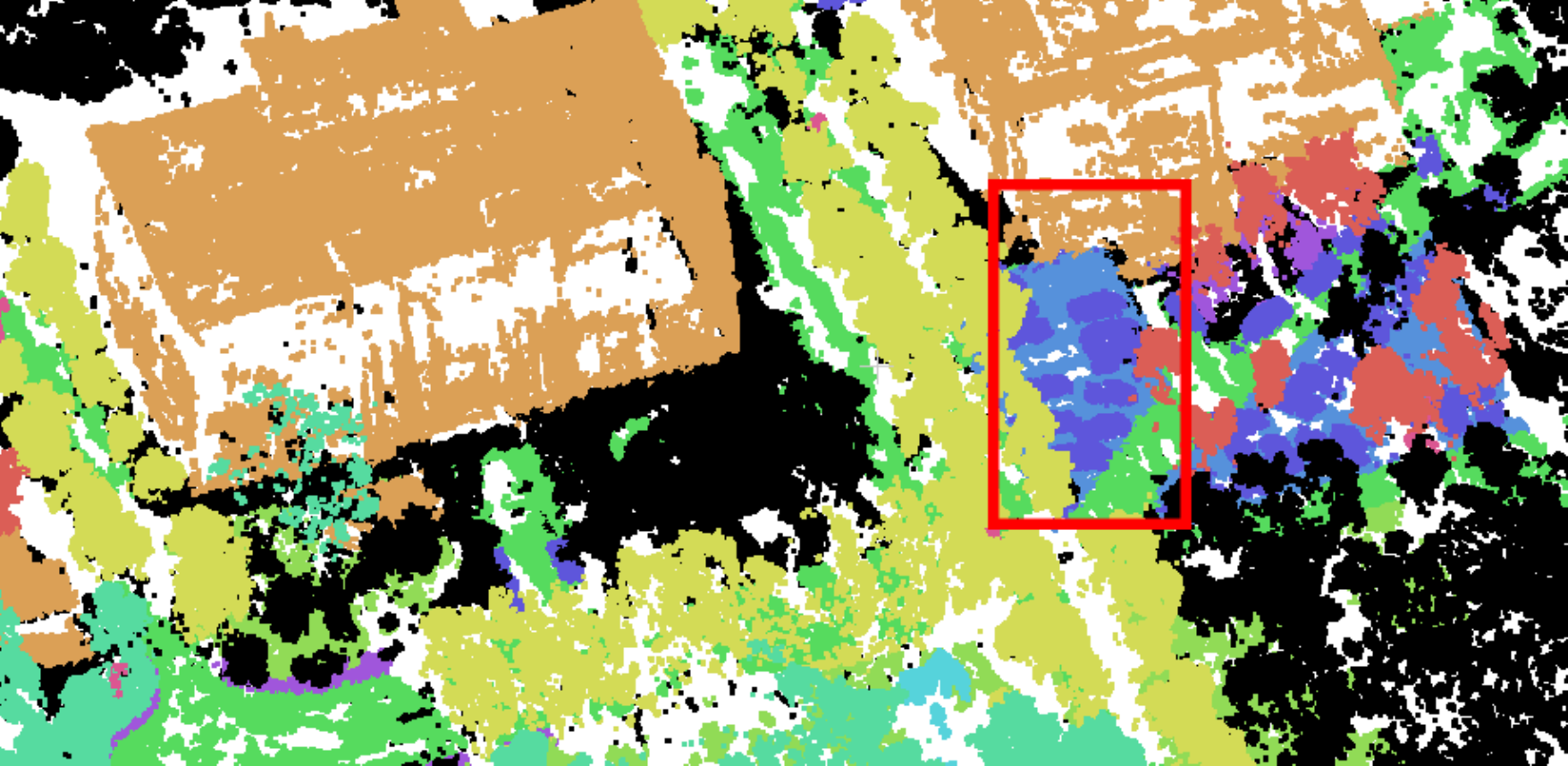}
            \end{minipage}
        }
    \end{minipage}
    \hspace{0.02\textwidth}
    \begin{minipage}[b]{0.30\textwidth}
        \centering
        \subfloat[]{
            \begin{minipage}[t]{0.95\textwidth}
                \includegraphics[width=\linewidth]{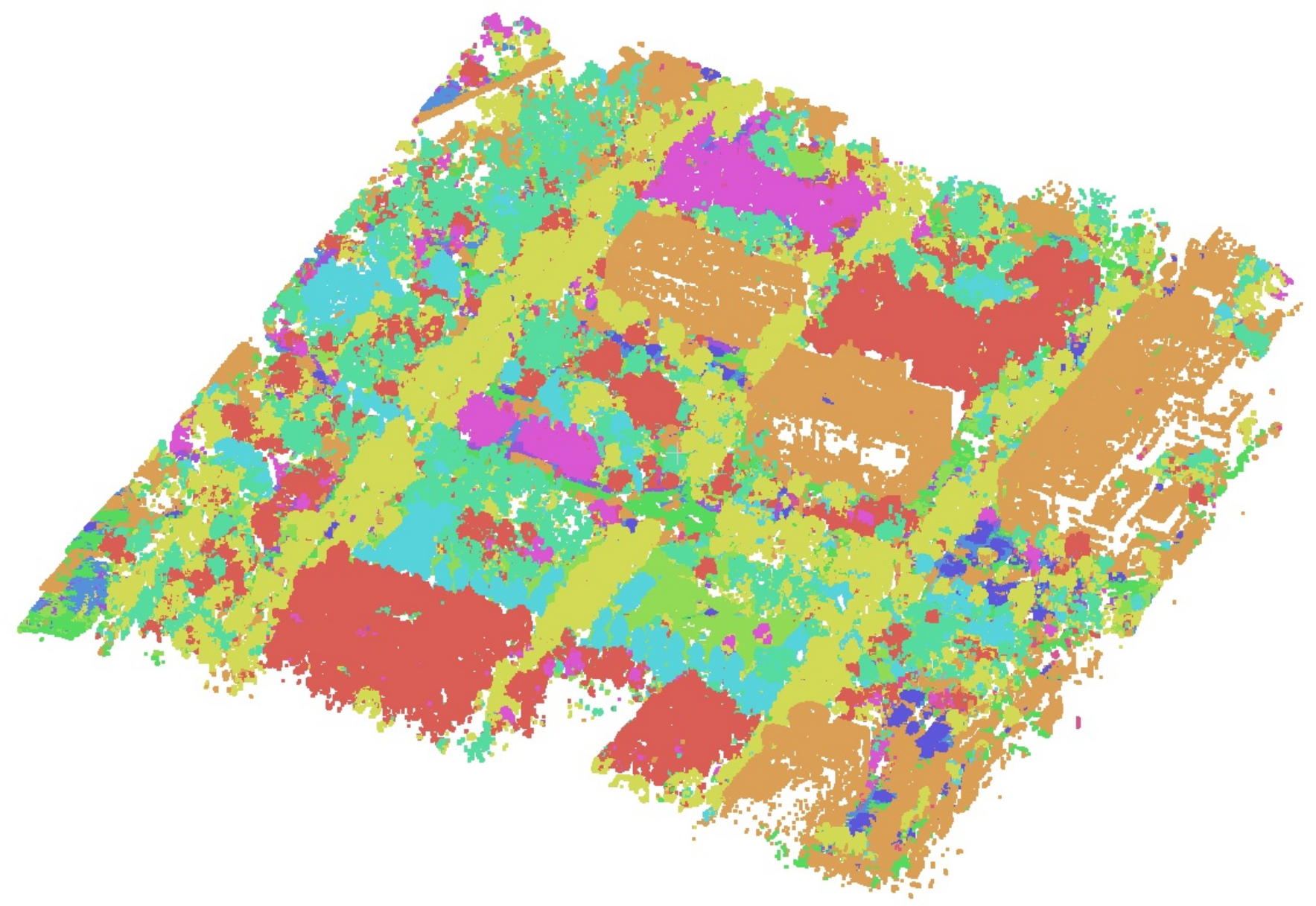}
                \includegraphics[width=\linewidth]{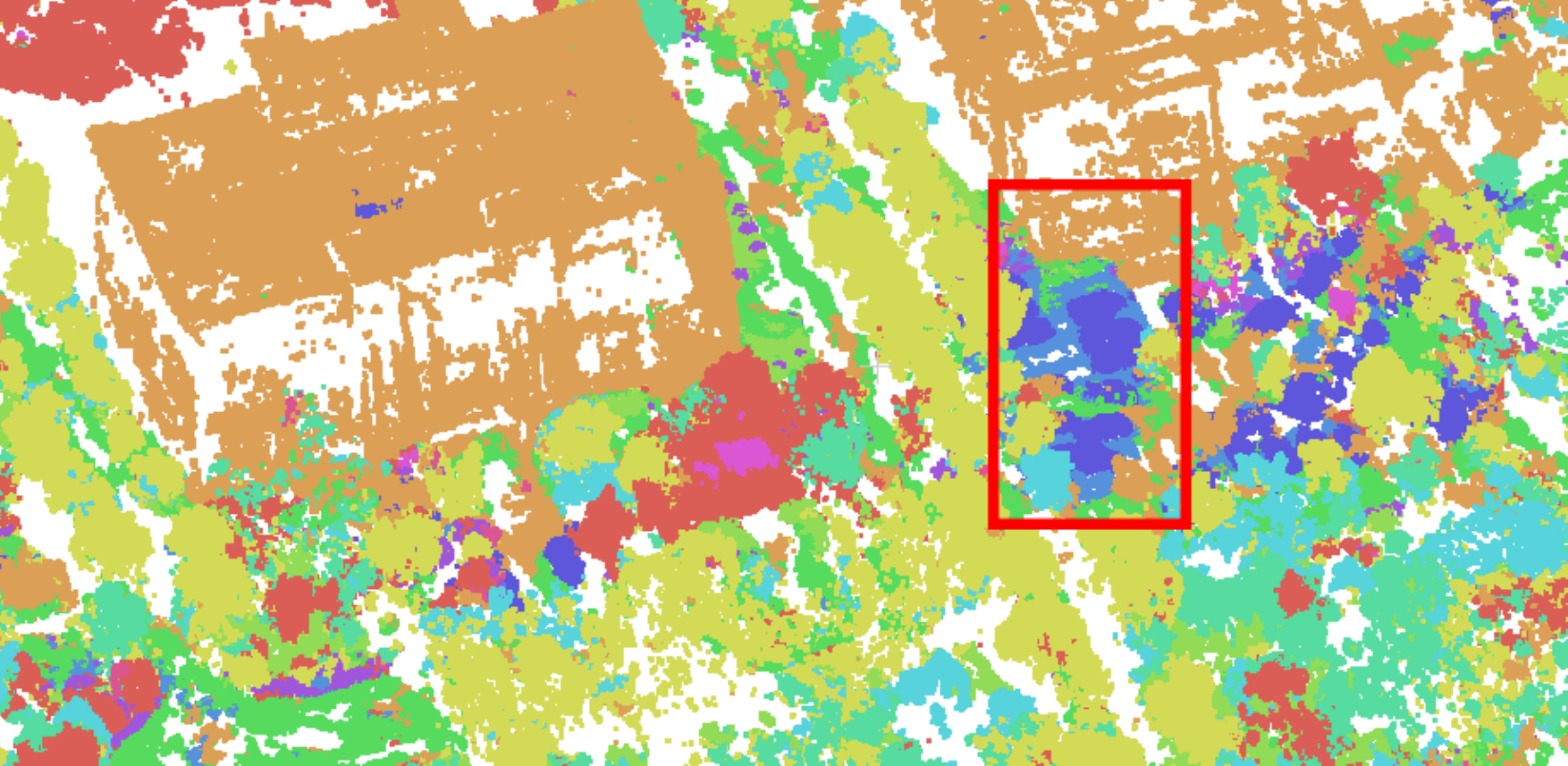}
            \end{minipage}
        }
    \end{minipage}
    \hspace{0.02\textwidth}
    \begin{minipage}[b]{0.30\textwidth}
        \centering
        \subfloat[]{
            \begin{minipage}[t]{0.95\textwidth}
                \includegraphics[width=\linewidth]{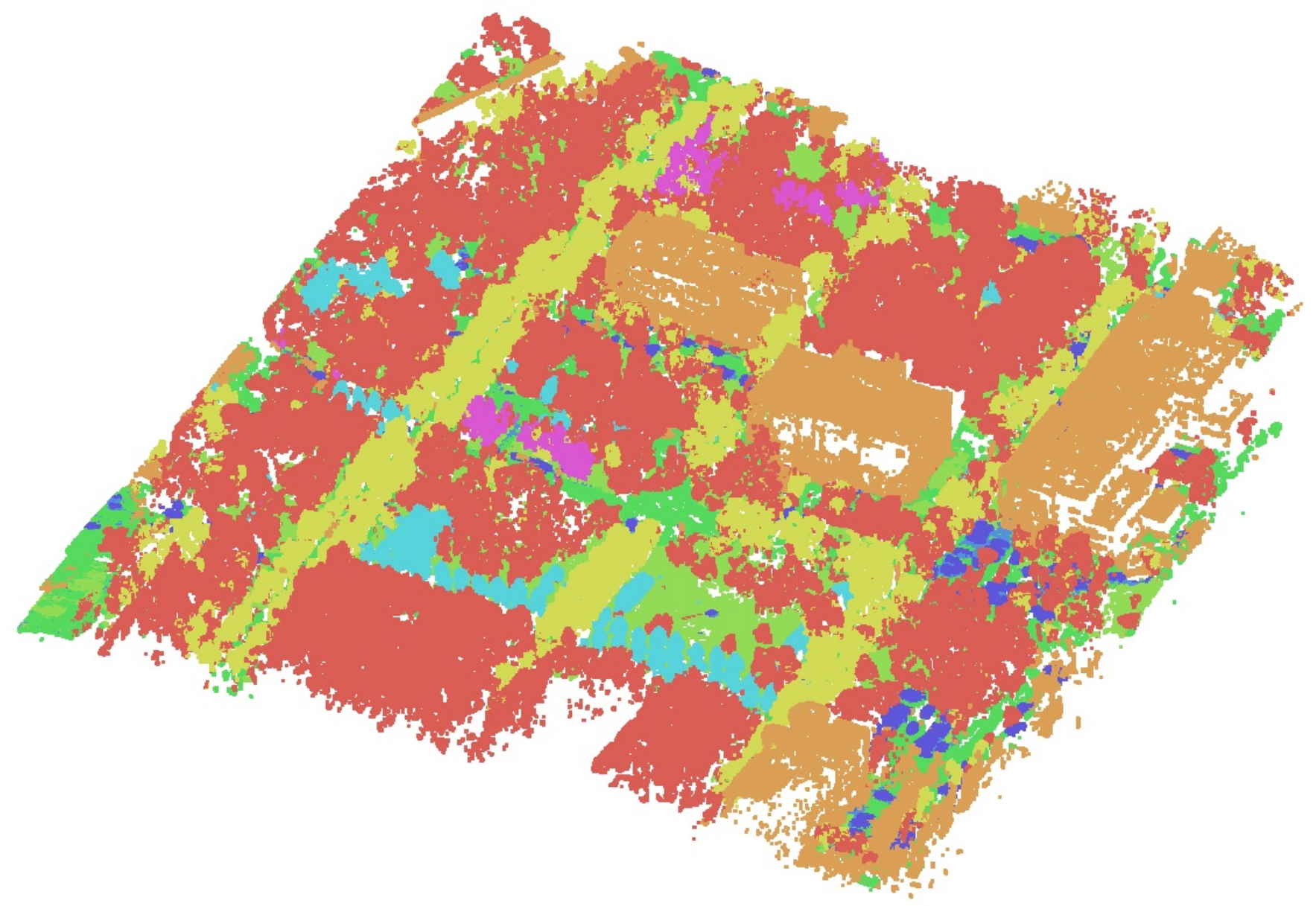}
                \includegraphics[width=\linewidth]{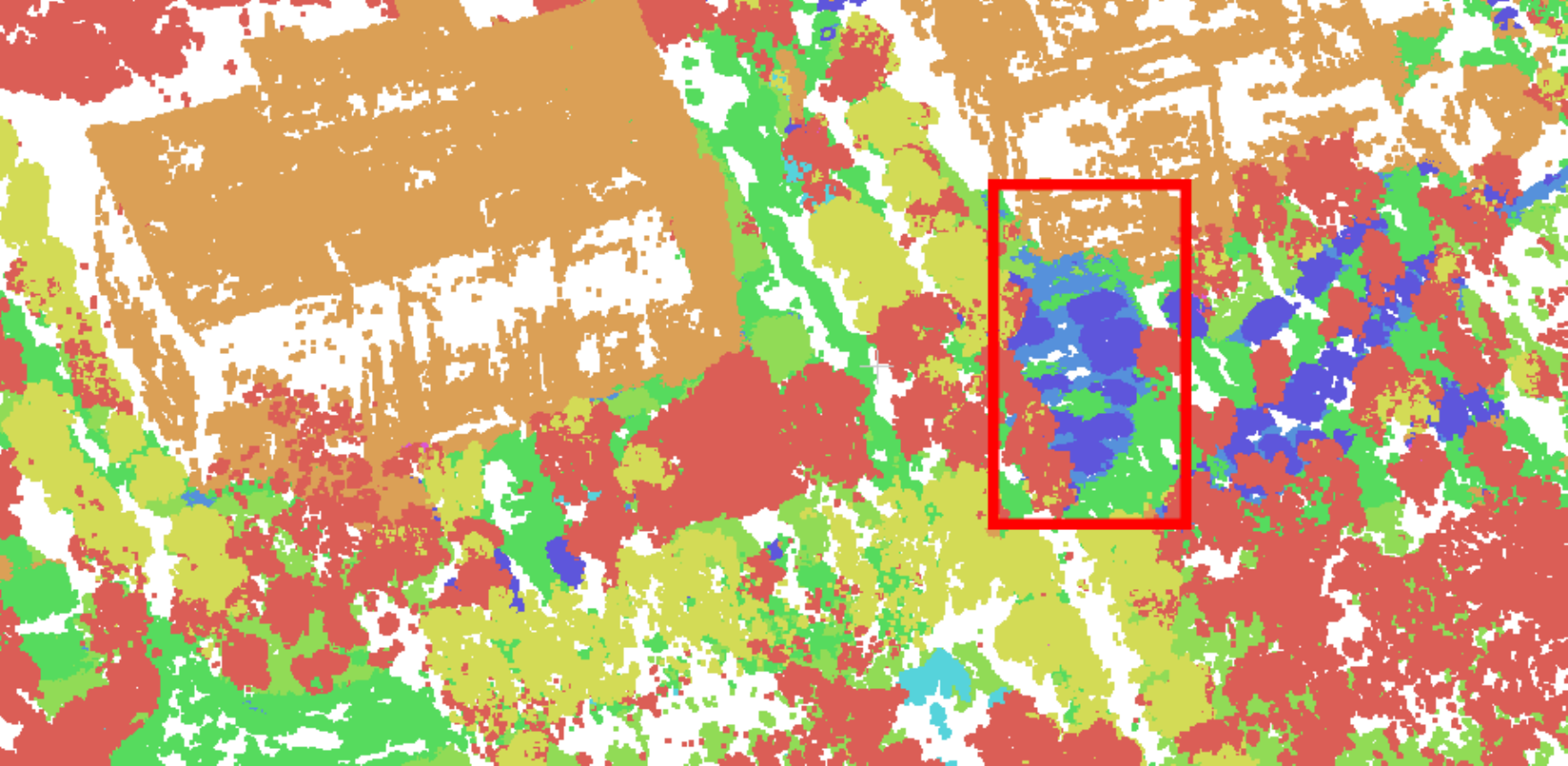}
            \end{minipage}
        }
    \end{minipage}
    \hspace{0.02\textwidth}
    \begin{minipage}[b]{0.30\textwidth}
        \centering
        \subfloat[]{
            \begin{minipage}[t]{0.95\textwidth}
                \includegraphics[width=\linewidth]{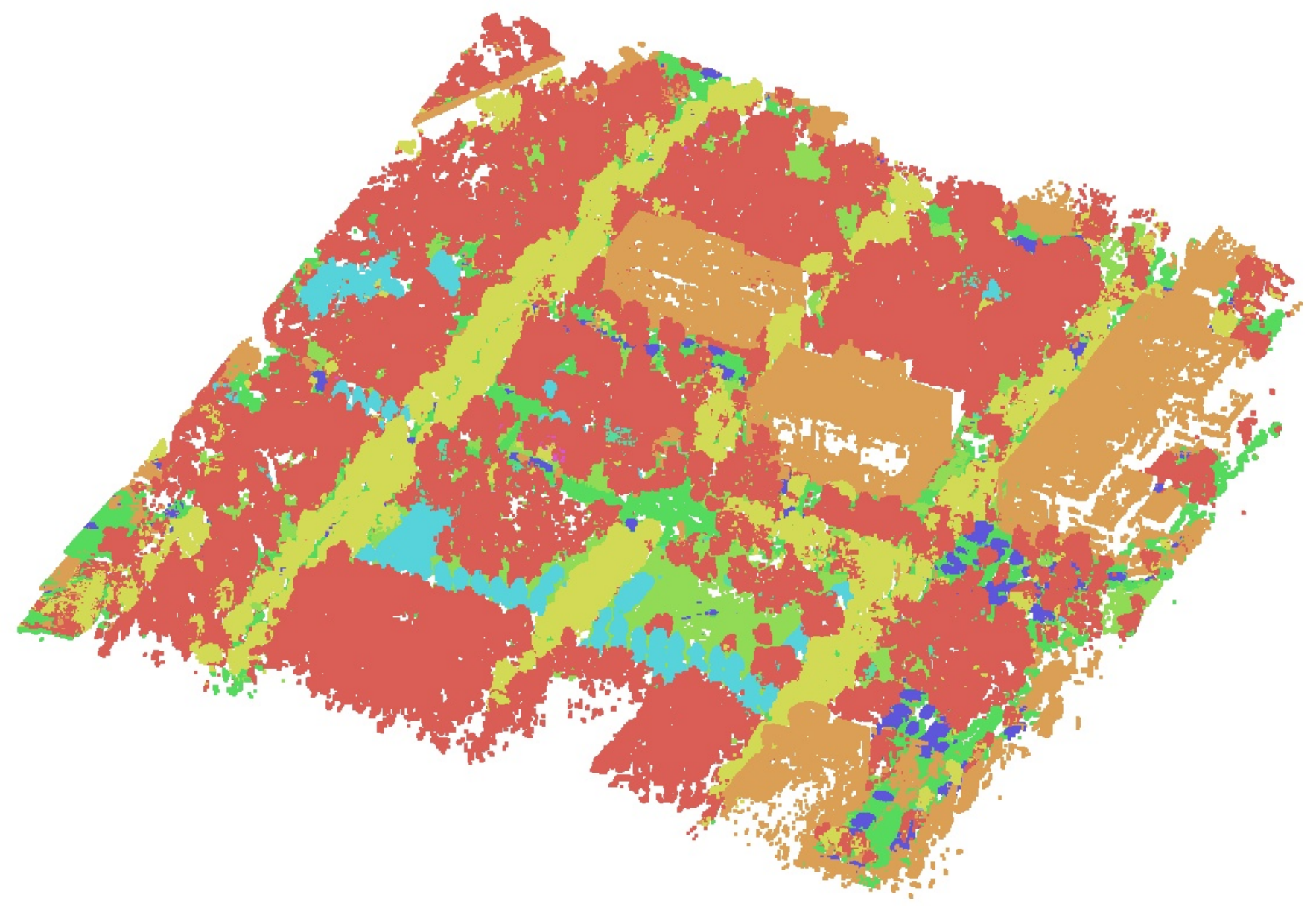}
                \includegraphics[width=\linewidth]{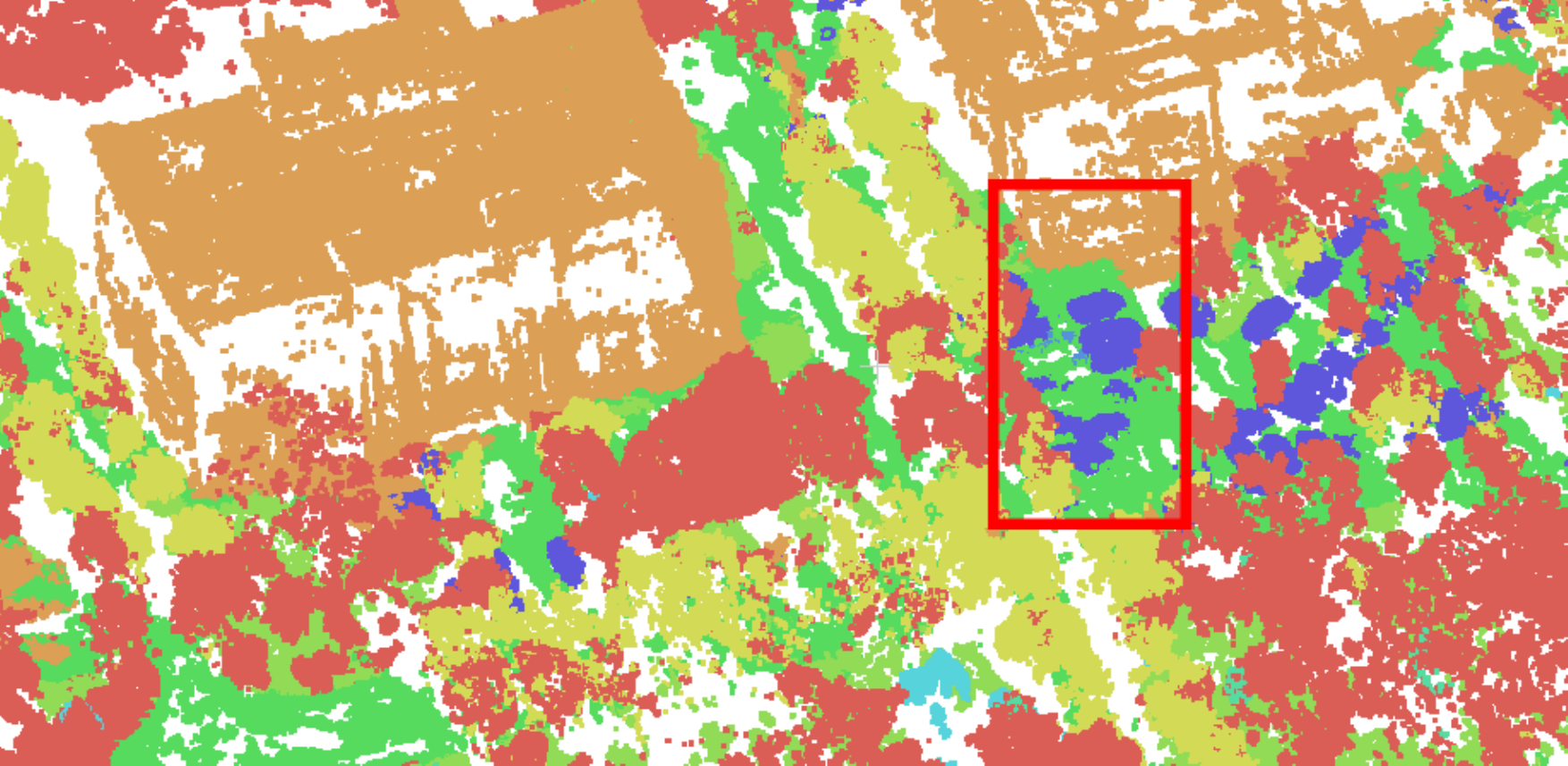}
            \end{minipage}
        }
    \end{minipage}
    \hspace{0.02\textwidth}
    \begin{minipage}[b]{0.30\textwidth}
        \centering
        \subfloat[]{
            \begin{minipage}[t]{0.95\textwidth}
                \includegraphics[width=\linewidth]{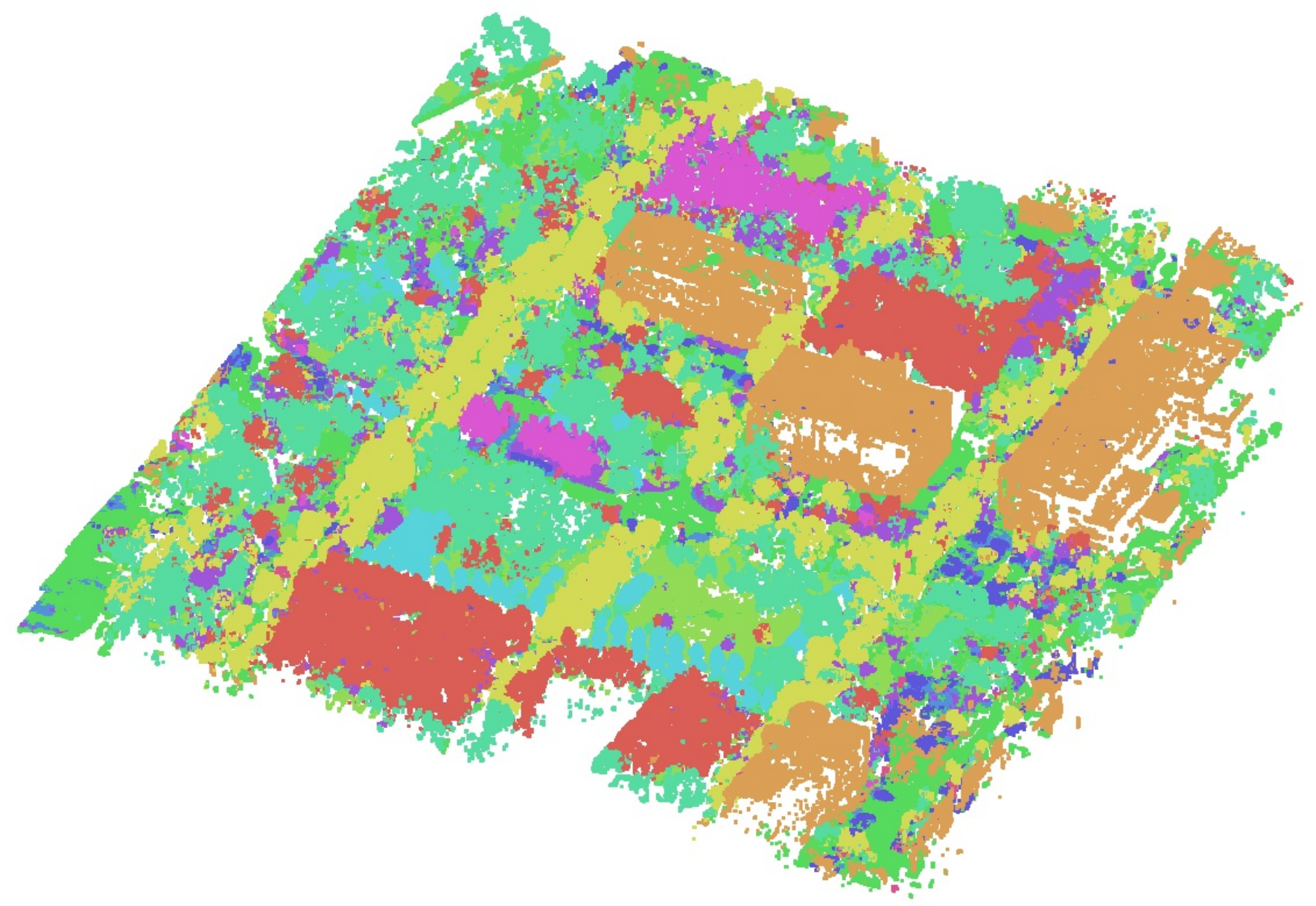}
                \includegraphics[width=\linewidth]{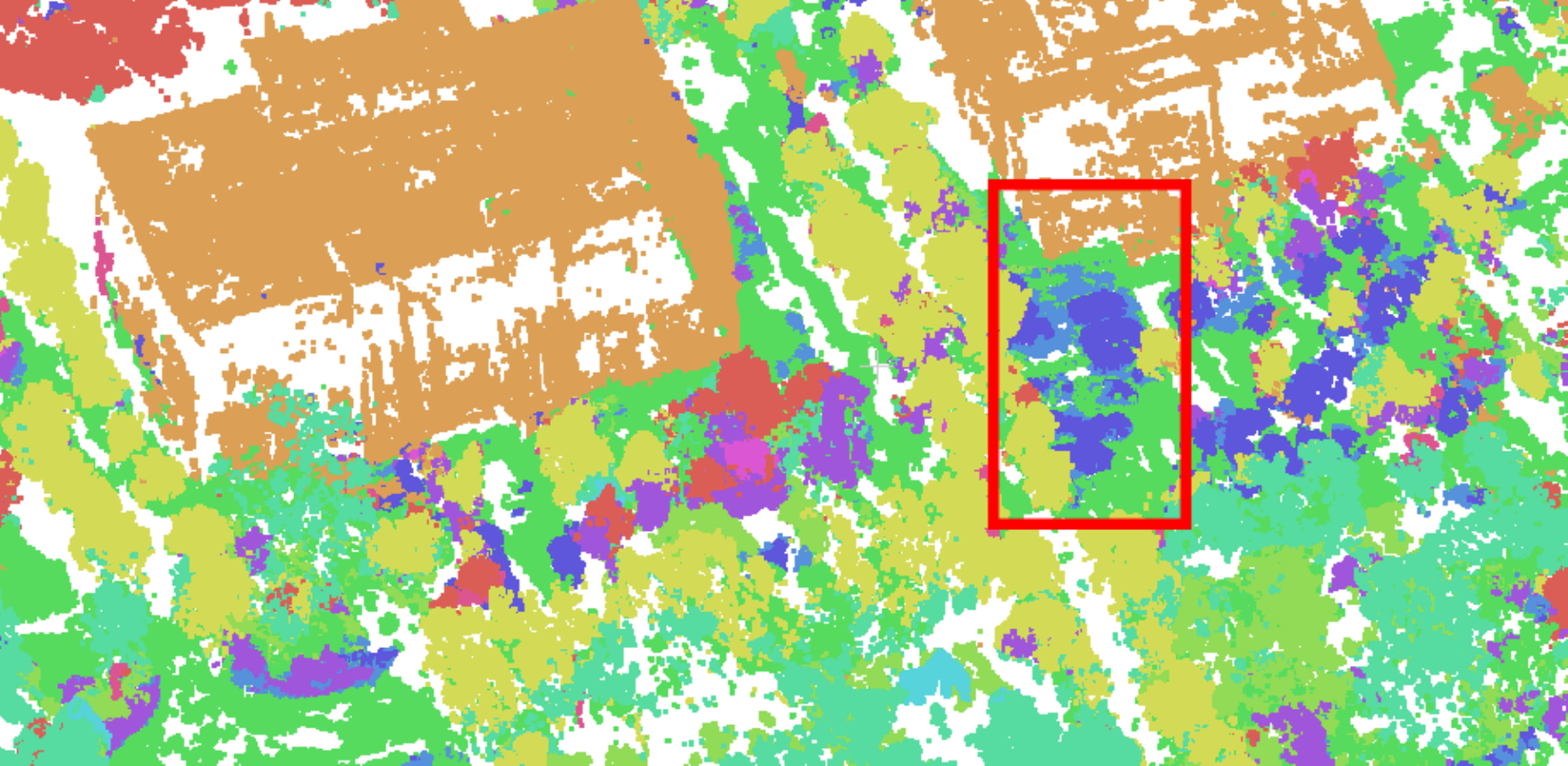}
            \end{minipage}
        }
    \end{minipage}
    \hspace{0.02\textwidth}
    \begin{minipage}[b]{0.30\textwidth}
        \centering
        \subfloat[]{
            \begin{minipage}[t]{0.95\textwidth}
                \includegraphics[width=\linewidth]{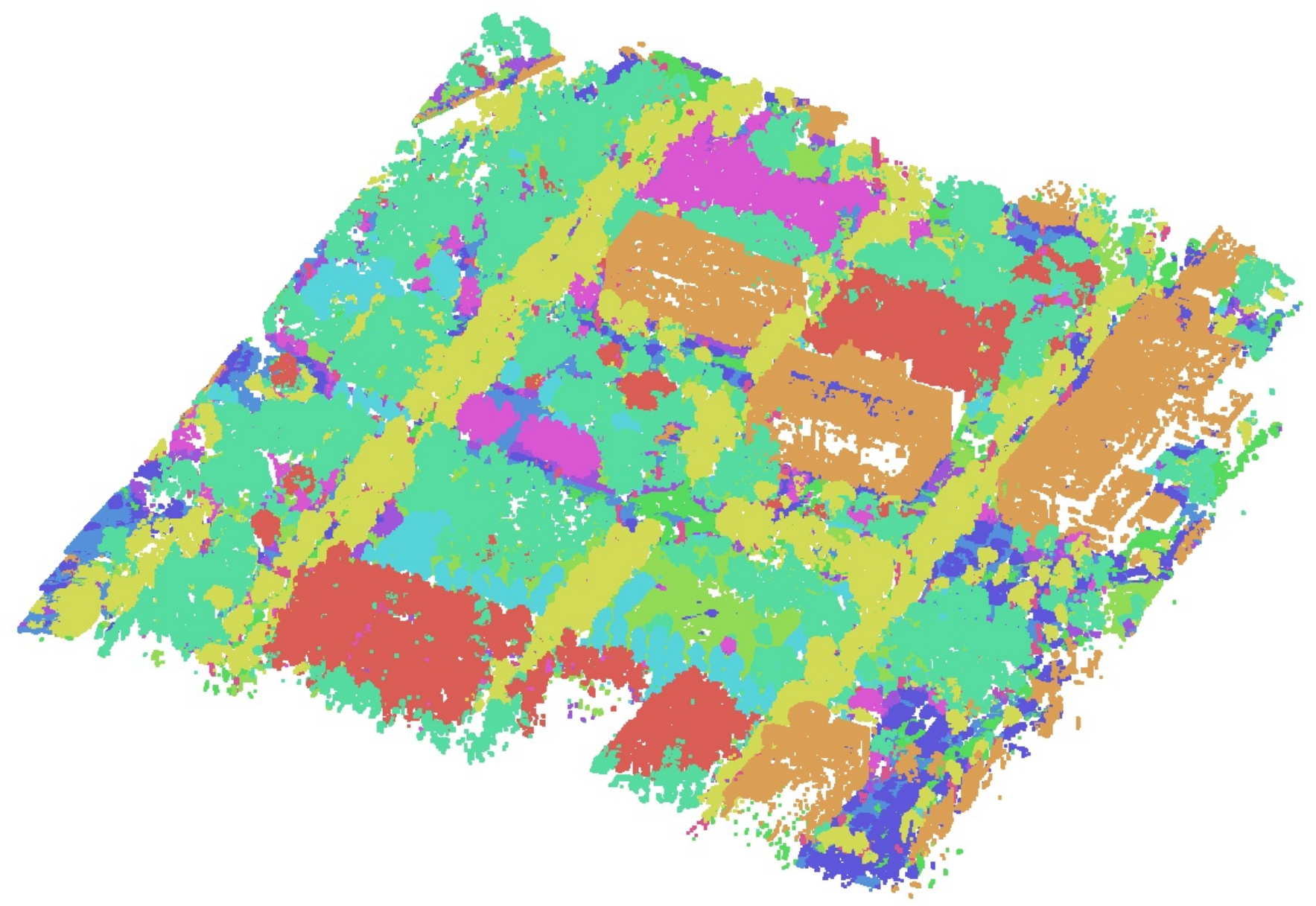}
                \includegraphics[width=\linewidth]{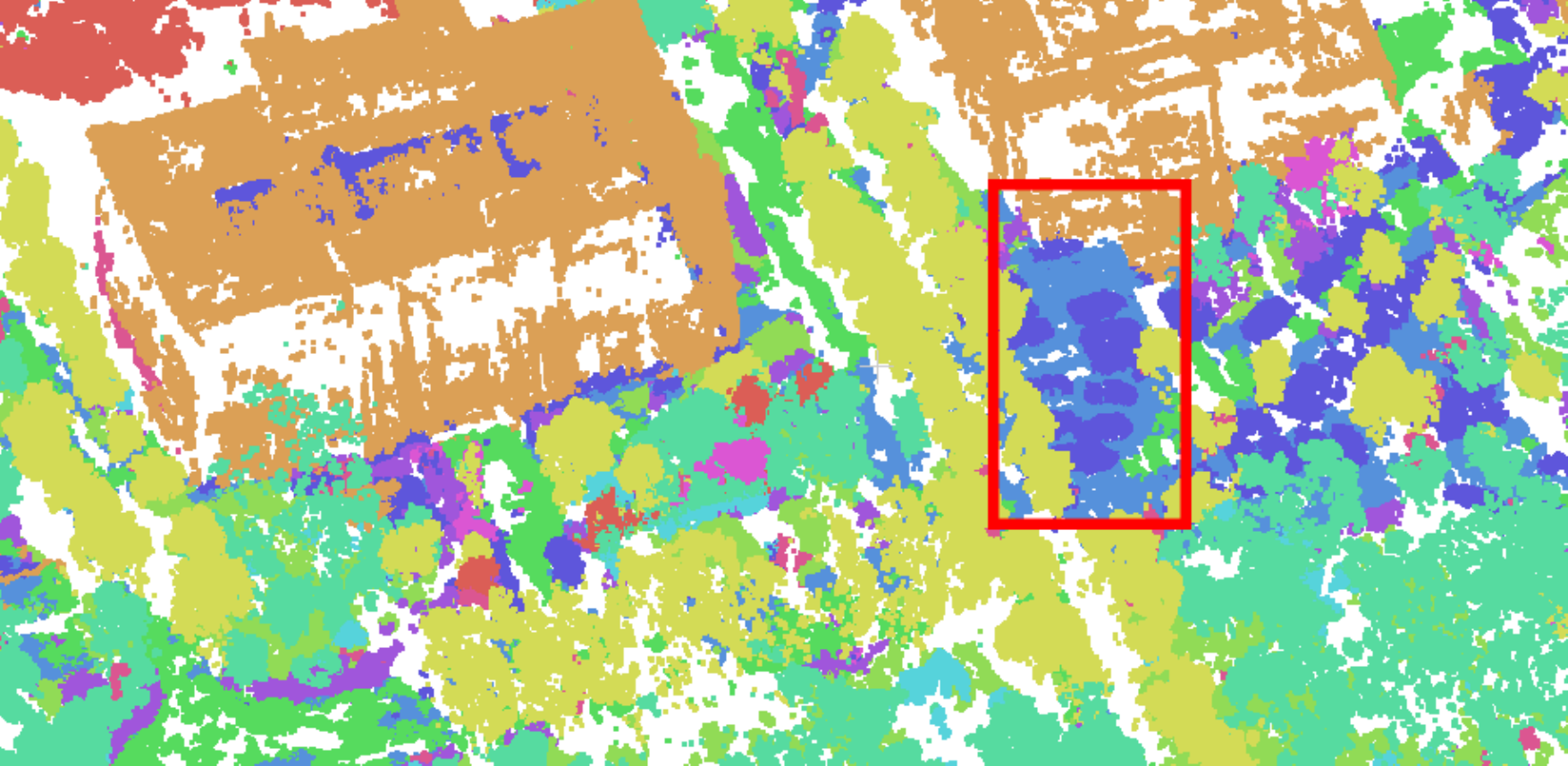}
            \end{minipage}
        }
    \end{minipage}
    \begin{minipage}[b]{0.98\textwidth}
        \centering
        \includegraphics[width=\linewidth]{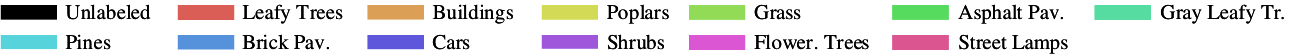}
    \end{minipage}
    \caption{Full classification maps on the HIT dataset obtained by (a) Ground truth, (b) RandLA-Net, (c) SCFNet, (d) CGANet, (e) Hu et al, (f) Ours.}
    \label{fig6}
    \end{figure*}

\begin{figure*}[!t]
    \centering
    \begin{minipage}[b]{0.30\textwidth}
        \centering
        \subfloat[]{
            \begin{minipage}[t]{0.95\textwidth}
                \includegraphics[width=\linewidth]{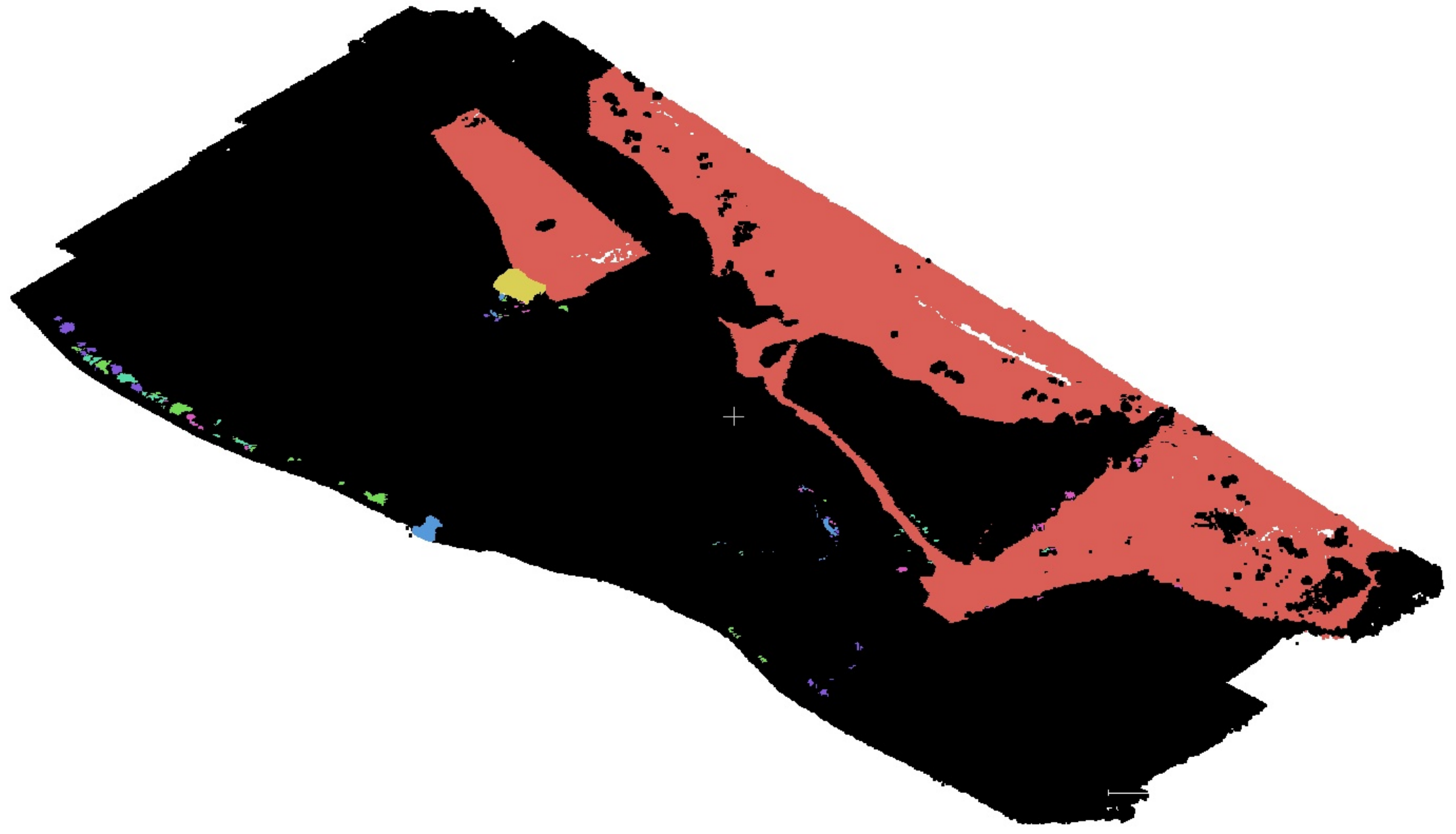}
                \includegraphics[width=\linewidth]{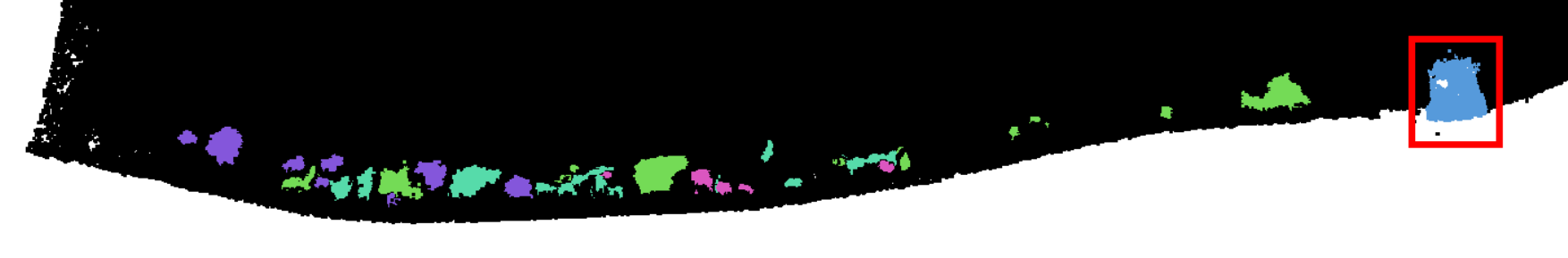}
            \end{minipage}
        }
    \end{minipage}
    \hspace{0.02\textwidth}
    \begin{minipage}[b]{0.30\textwidth}
        \centering
        \subfloat[]{
            \begin{minipage}[t]{0.95\textwidth}
                \includegraphics[width=\linewidth]{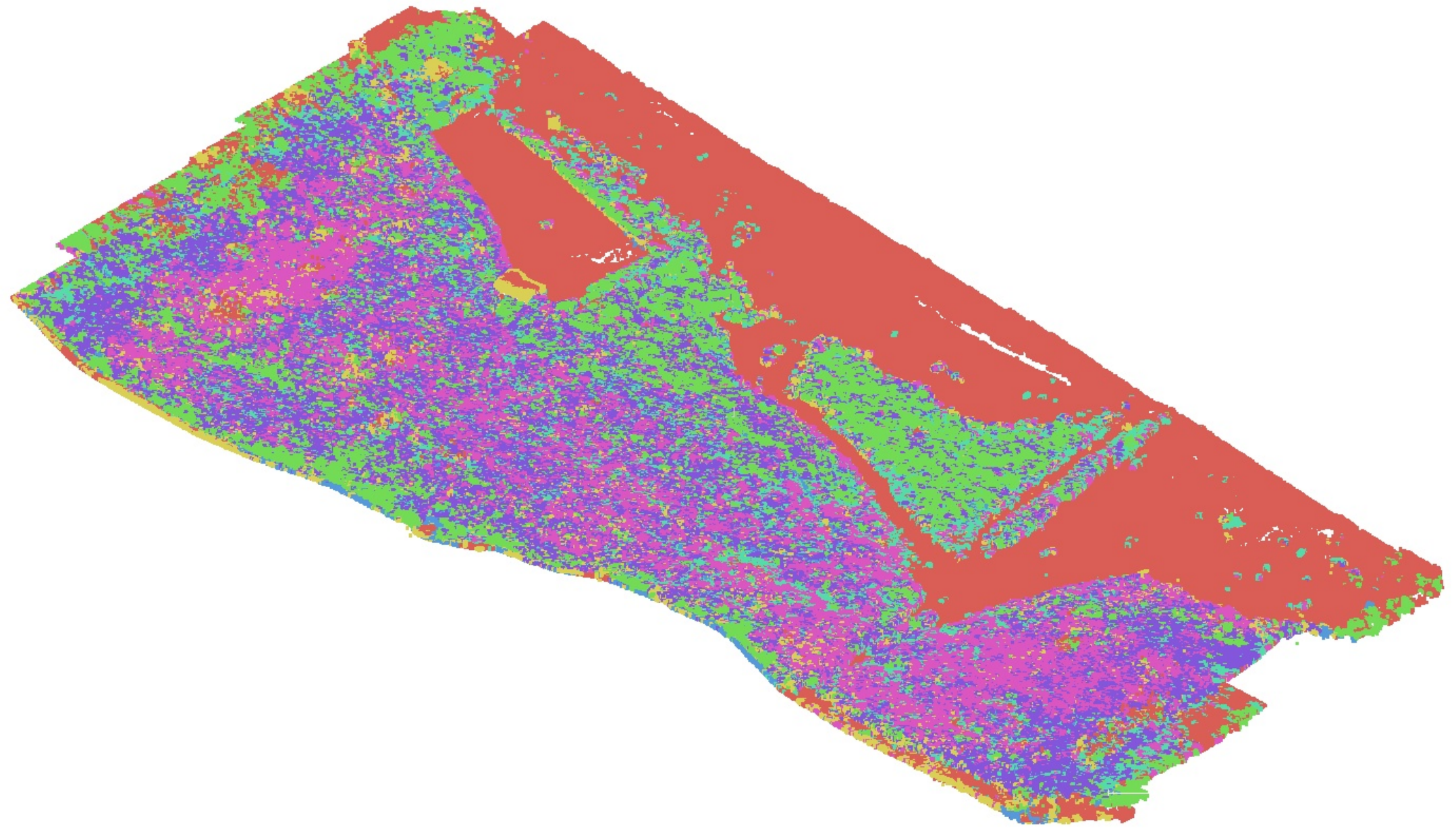}
                \includegraphics[width=\linewidth]{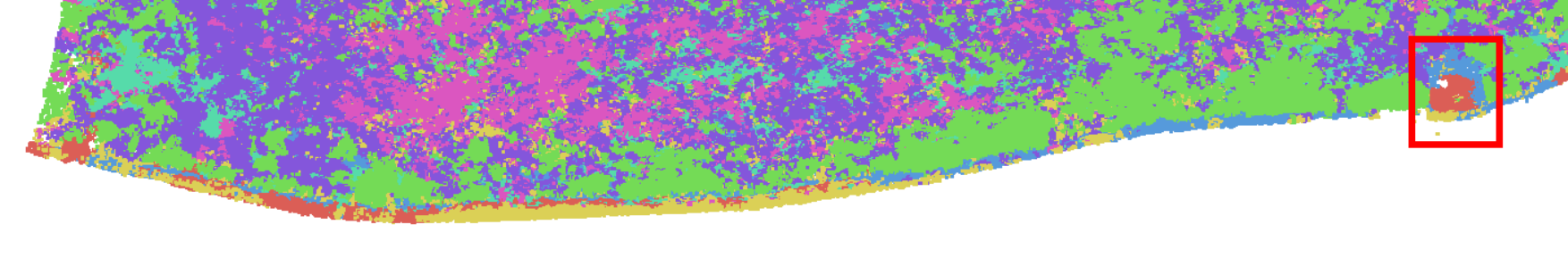}
            \end{minipage}
        }
    \end{minipage}
    \hspace{0.02\textwidth}
    \begin{minipage}[b]{0.30\textwidth}
        \centering
        \subfloat[]{
            \begin{minipage}[t]{0.95\textwidth}
                \includegraphics[width=\linewidth]{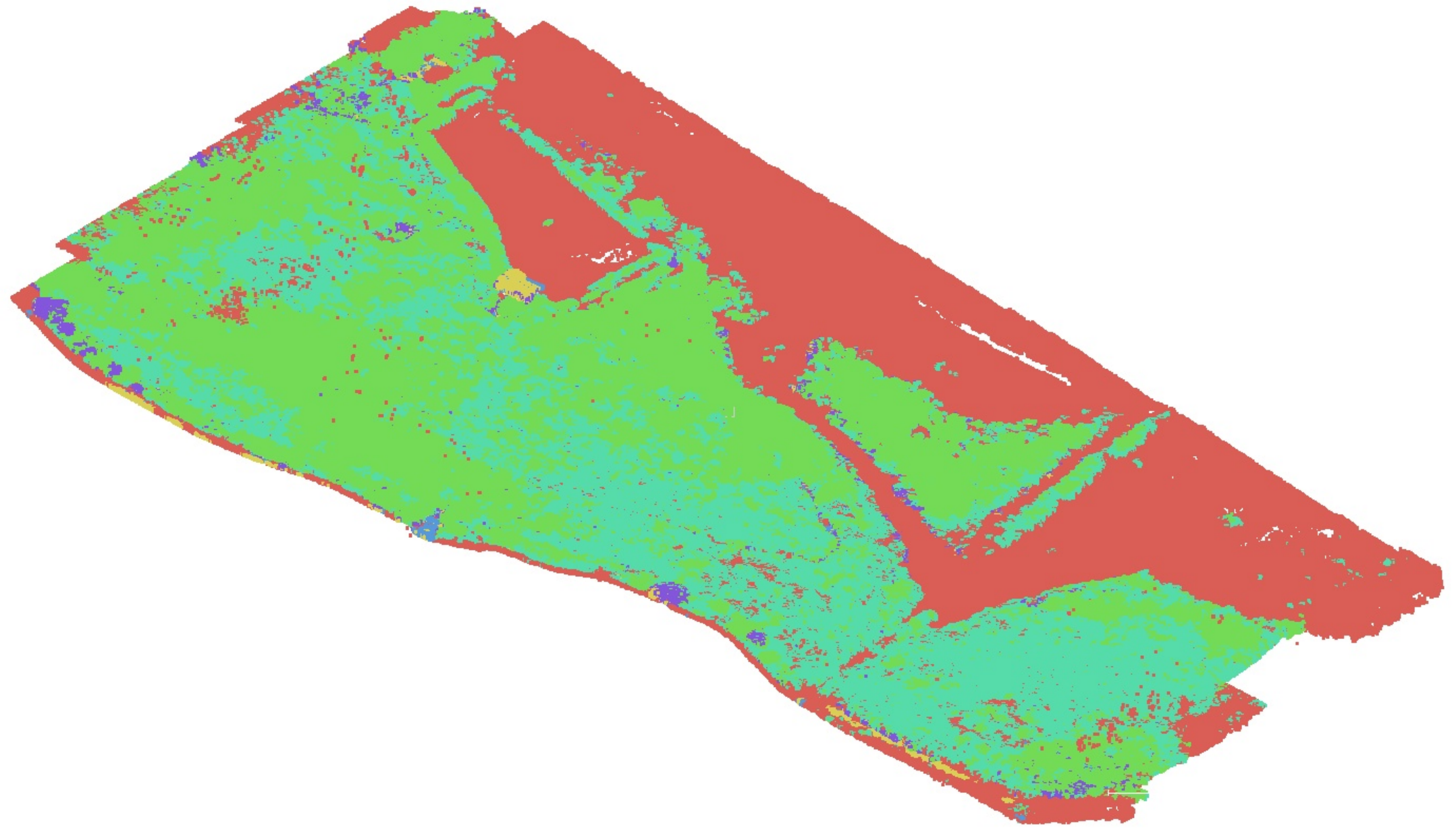}
                \includegraphics[width=\linewidth]{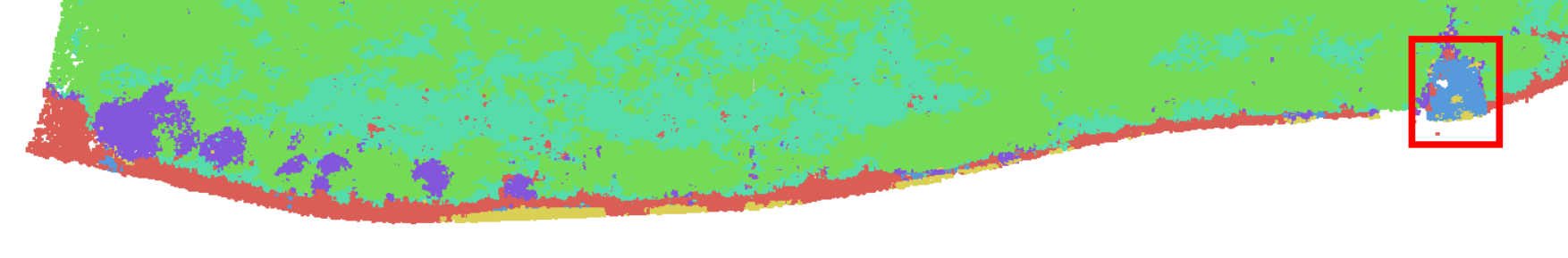}
            \end{minipage}
        }
    \end{minipage}
    \hspace{0.02\textwidth}
    \begin{minipage}[b]{0.30\textwidth}
        \centering
        \subfloat[]{
            \begin{minipage}[t]{0.95\textwidth}
                \includegraphics[width=\linewidth]{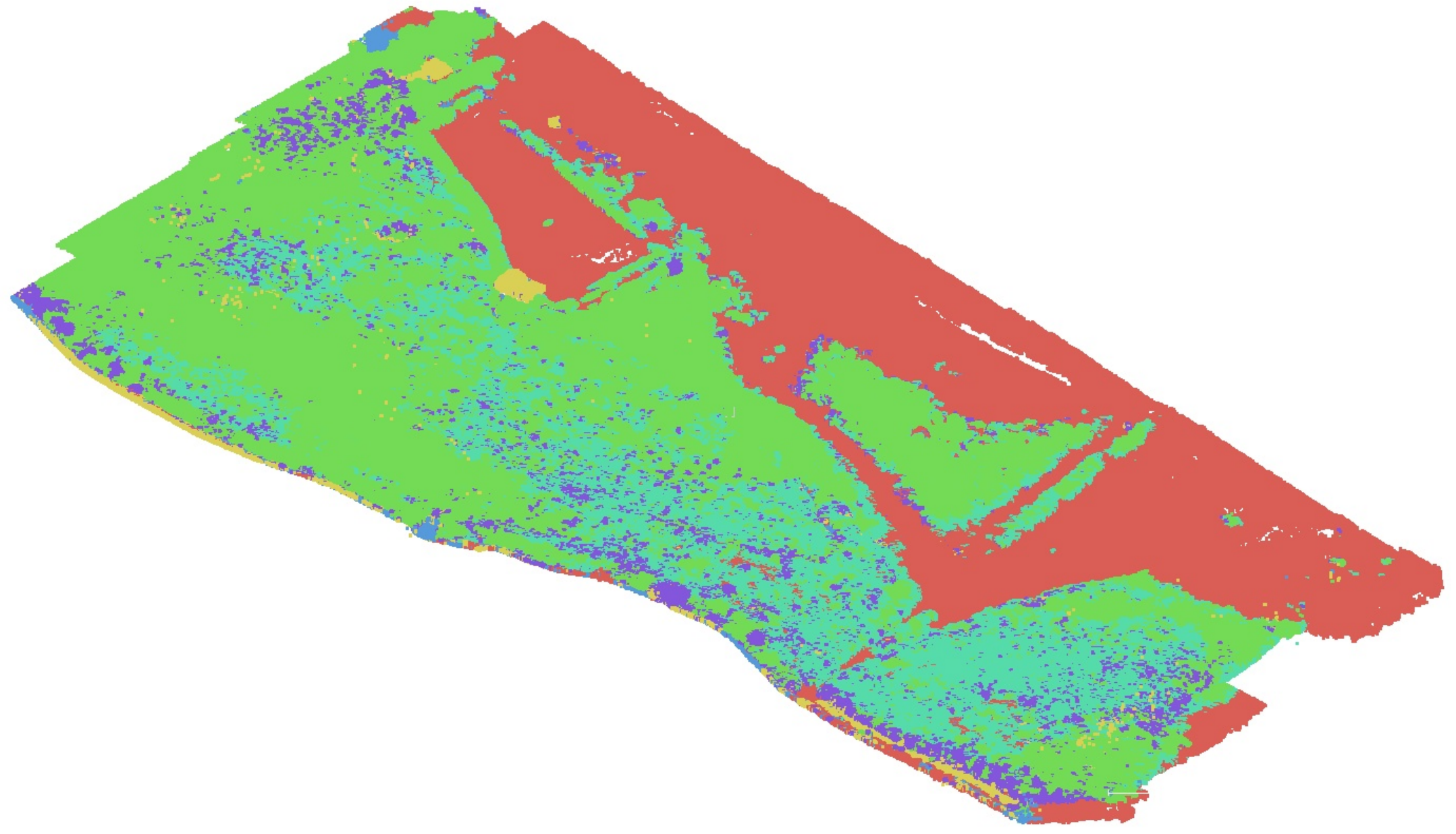}
                \includegraphics[width=\linewidth]{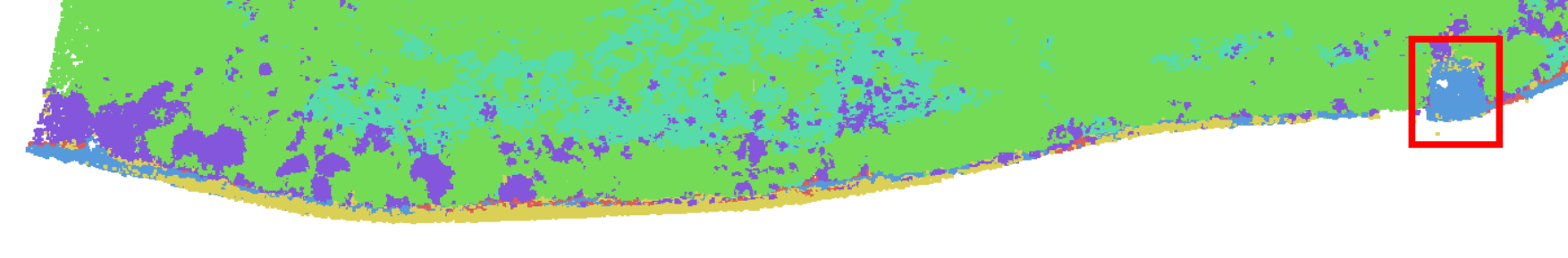}
            \end{minipage}
        }
    \end{minipage}
    \hspace{0.02\textwidth}
    \begin{minipage}[b]{0.30\textwidth}
        \centering
        \subfloat[]{
            \begin{minipage}[t]{0.95\textwidth}
                \includegraphics[width=\linewidth]{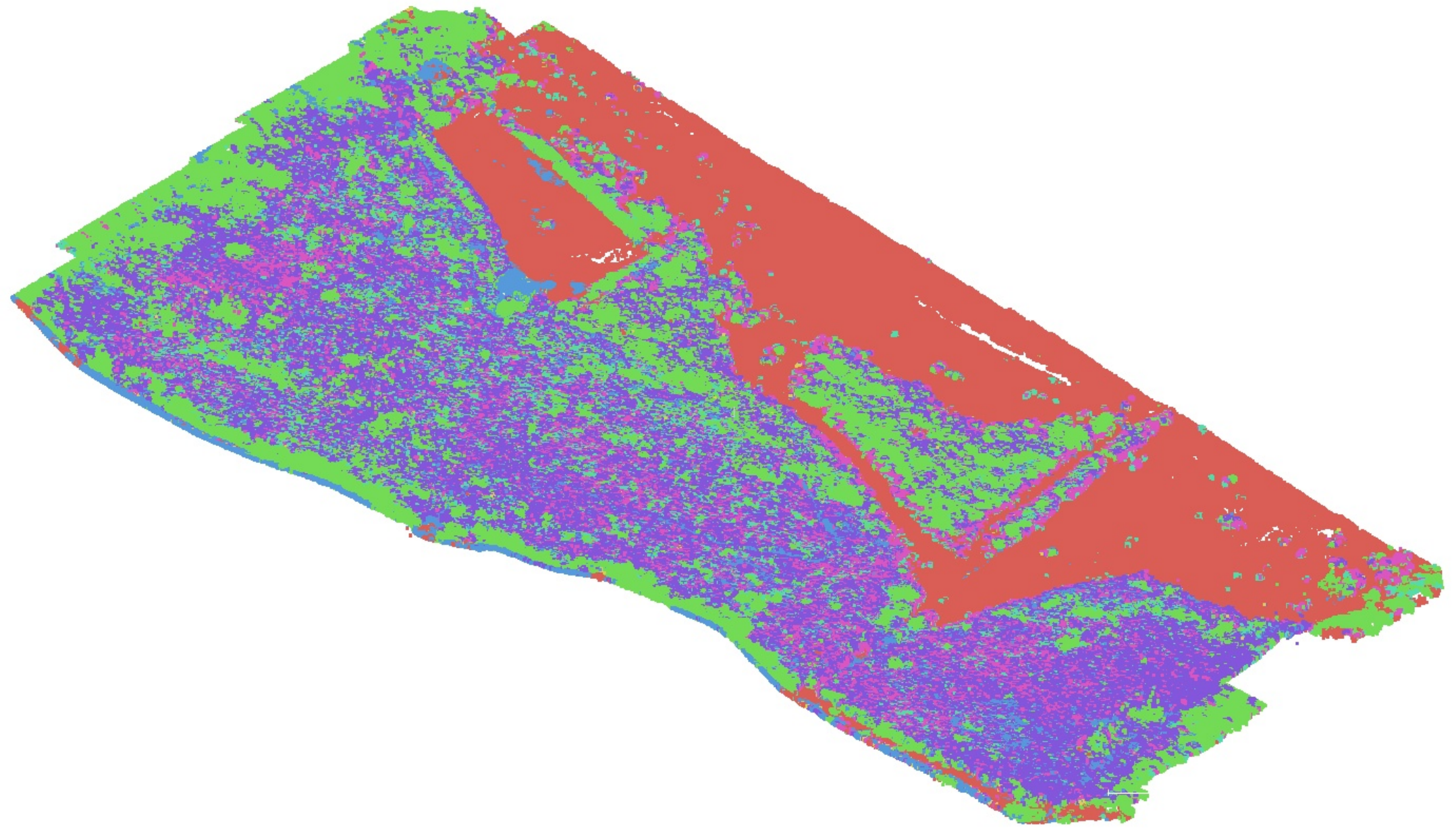}
                \includegraphics[width=\linewidth]{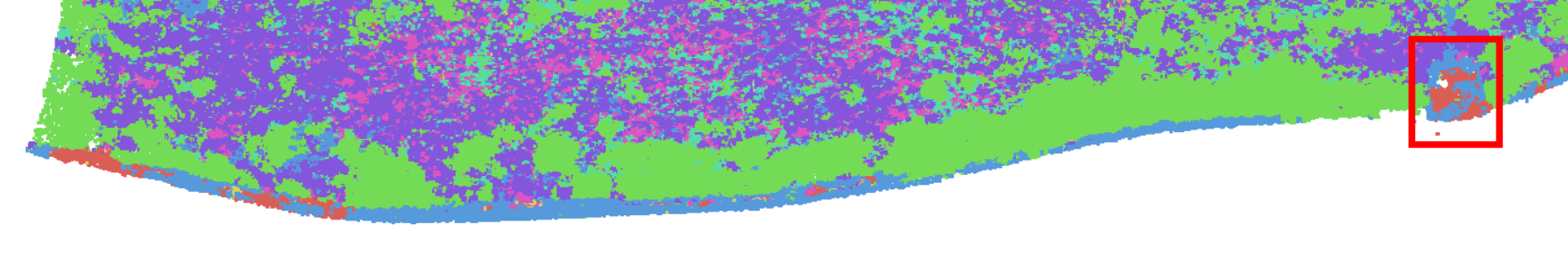}
            \end{minipage}
        }
    \end{minipage}
    \hspace{0.02\textwidth}
    \begin{minipage}[b]{0.30\textwidth}
        \centering
        \subfloat[]{
            \begin{minipage}[t]{0.95\textwidth}
                \includegraphics[width=\linewidth]{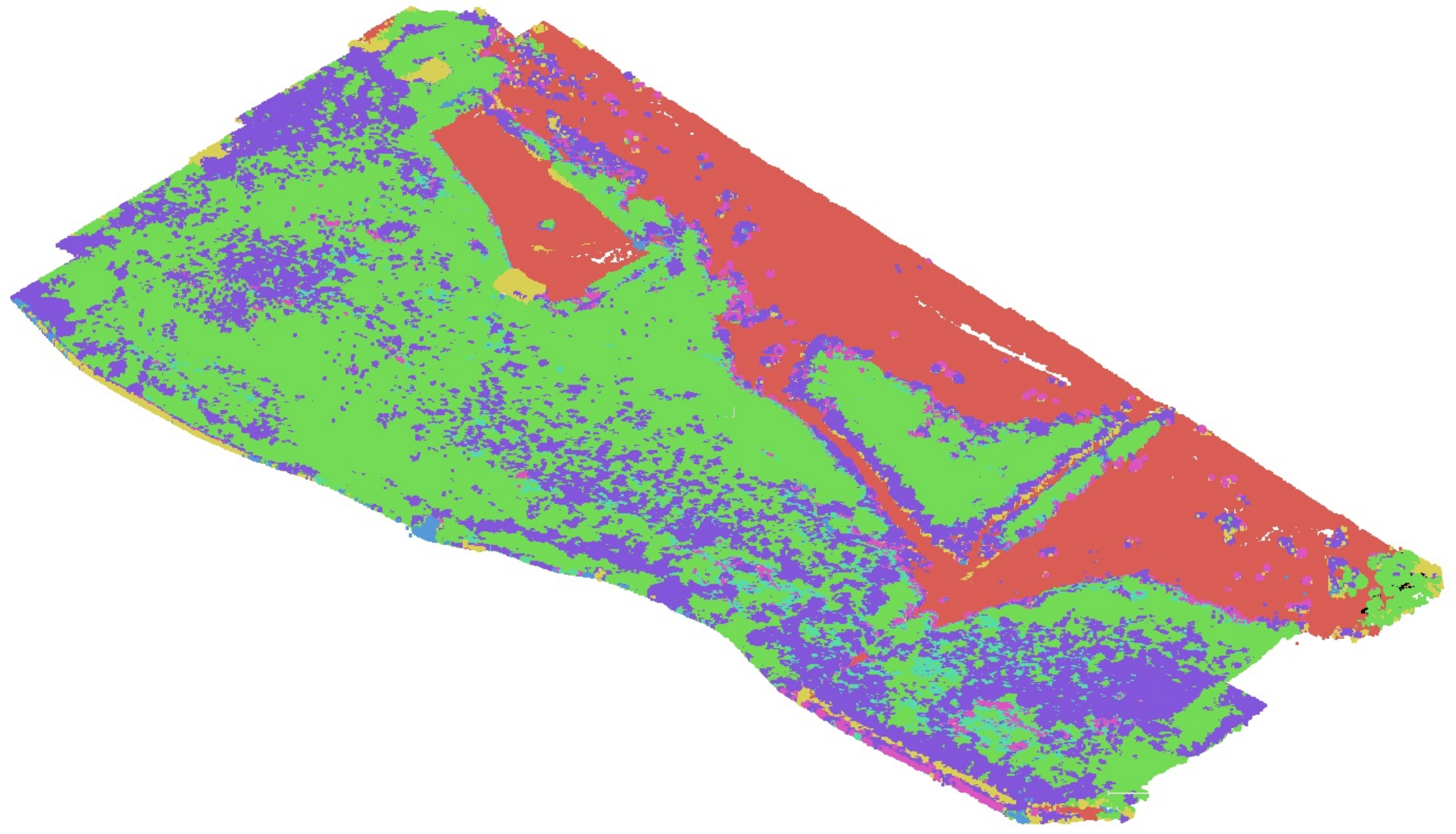}
                \includegraphics[width=\linewidth]{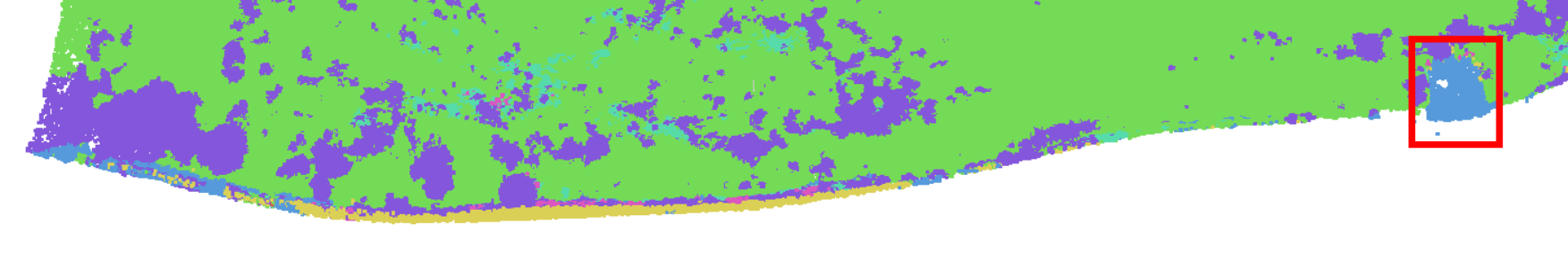}
            \end{minipage}
        }
    \end{minipage}
    \begin{minipage}[b]{0.98\textwidth}
        \centering
        \includegraphics[width=0.65\linewidth]{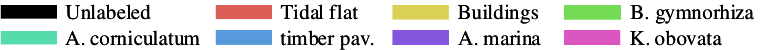}
    \end{minipage}
    \caption{Full classification maps on the ZJK dataset obtained by (a) Ground truth, (b) RandLA-Net, (c) SCFNet, (d) CGANet, (e) Hu et al, (f) Ours.}
    \label{fig7}
    \end{figure*}

\begin{figure*}[!t]
    \centering
    \begin{minipage}[b]{0.30\textwidth}
        \centering
        \subfloat[]{
            \begin{minipage}[t]{0.95\textwidth}
                \includegraphics[width=\linewidth]{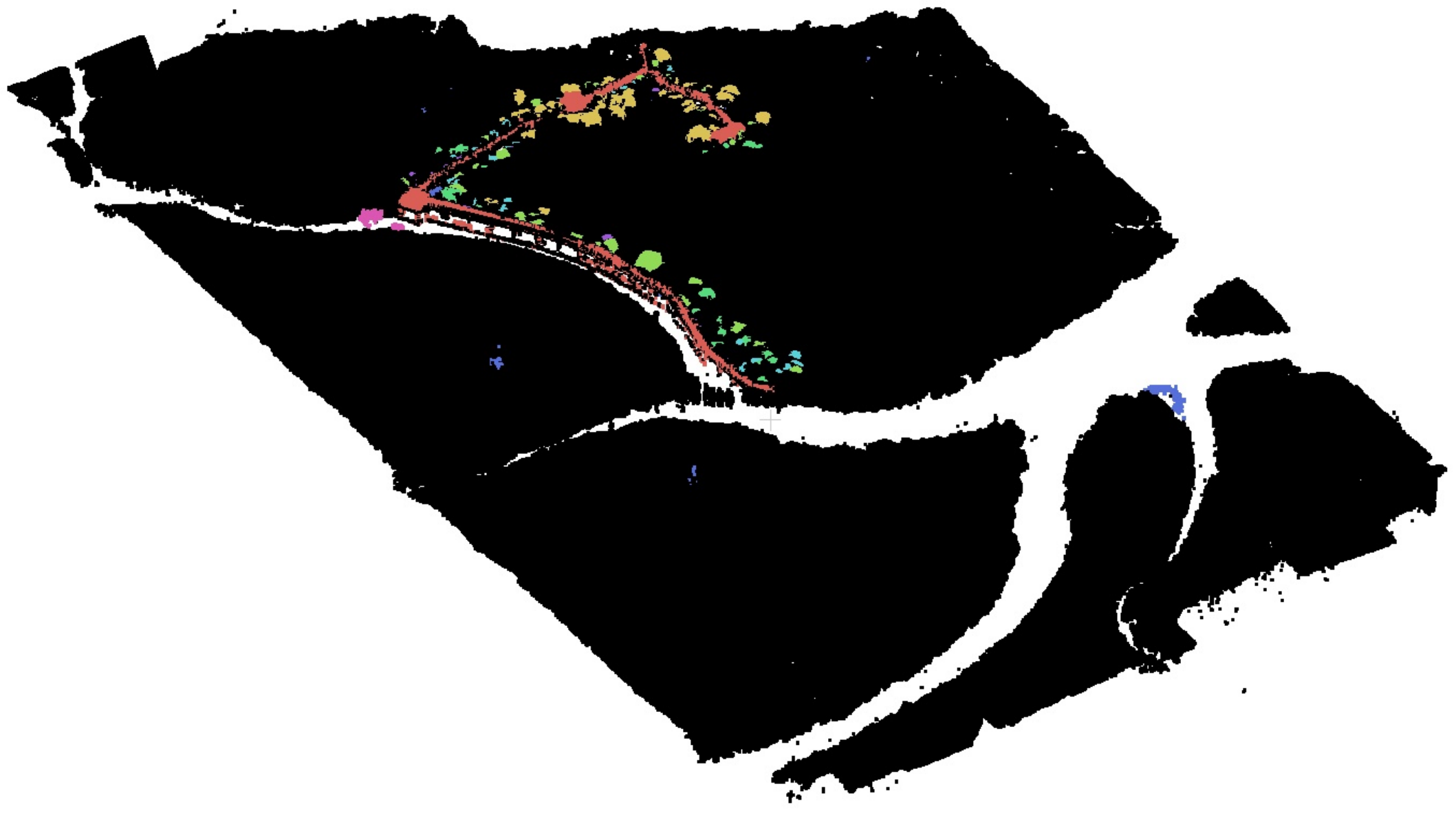}
                \includegraphics[width=\linewidth]{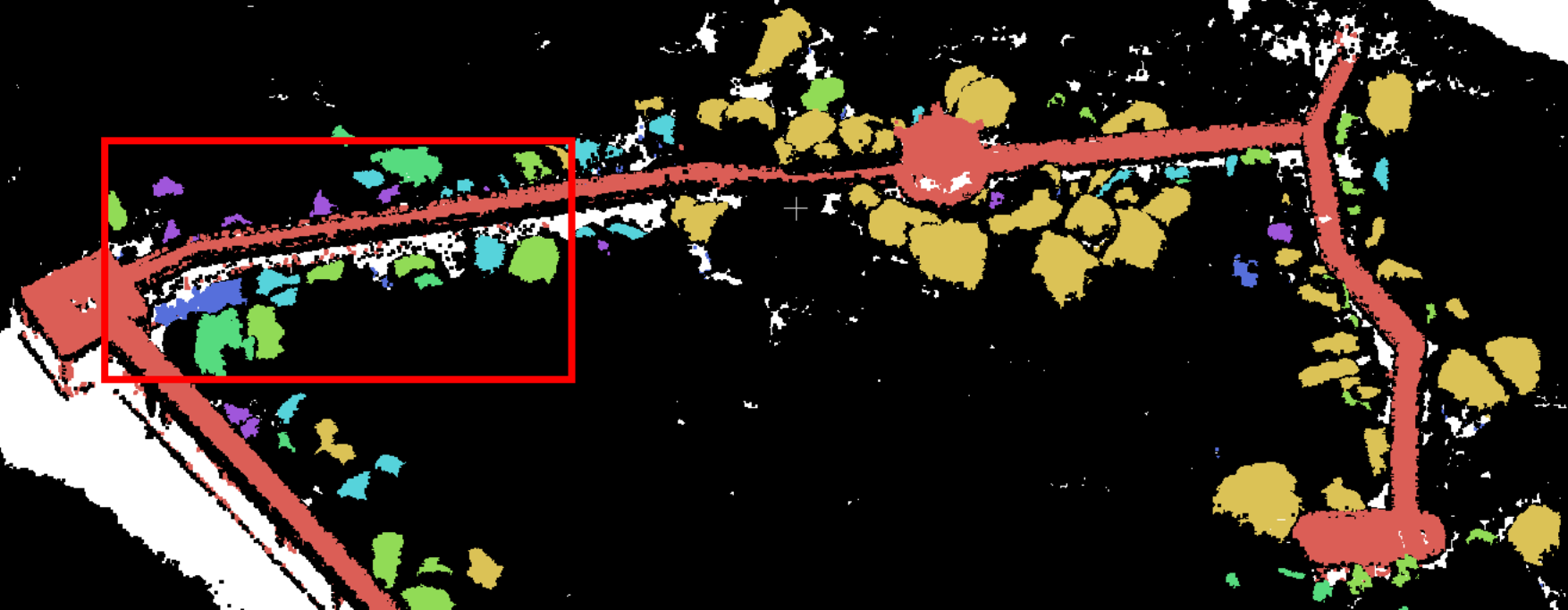}
            \end{minipage}
        }
    \end{minipage}
    \hspace{0.02\textwidth}
    \begin{minipage}[b]{0.30\textwidth}
        \centering
        \subfloat[]{
            \begin{minipage}[t]{0.95\textwidth}
                \includegraphics[width=\linewidth]{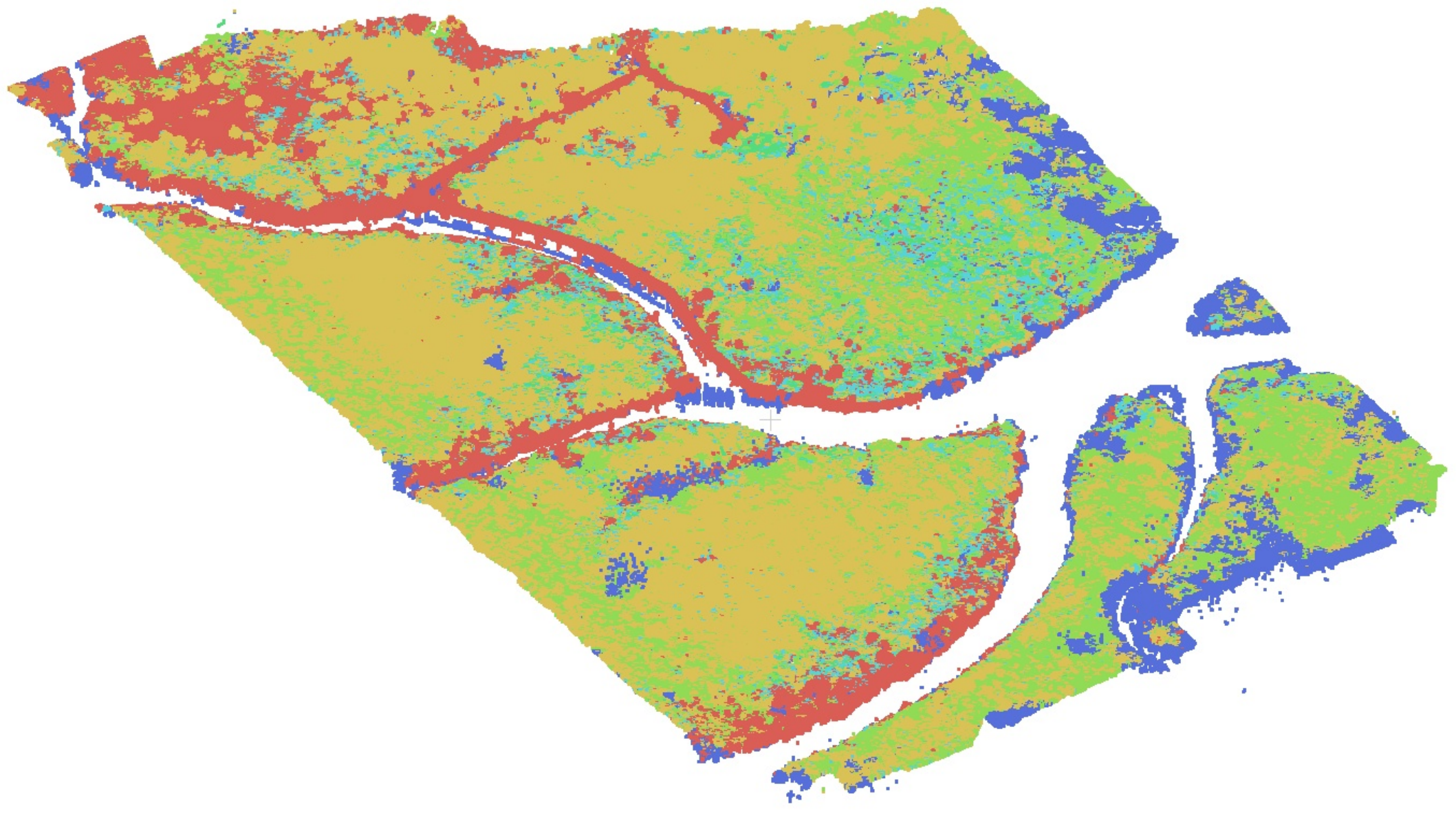}
                \includegraphics[width=\linewidth]{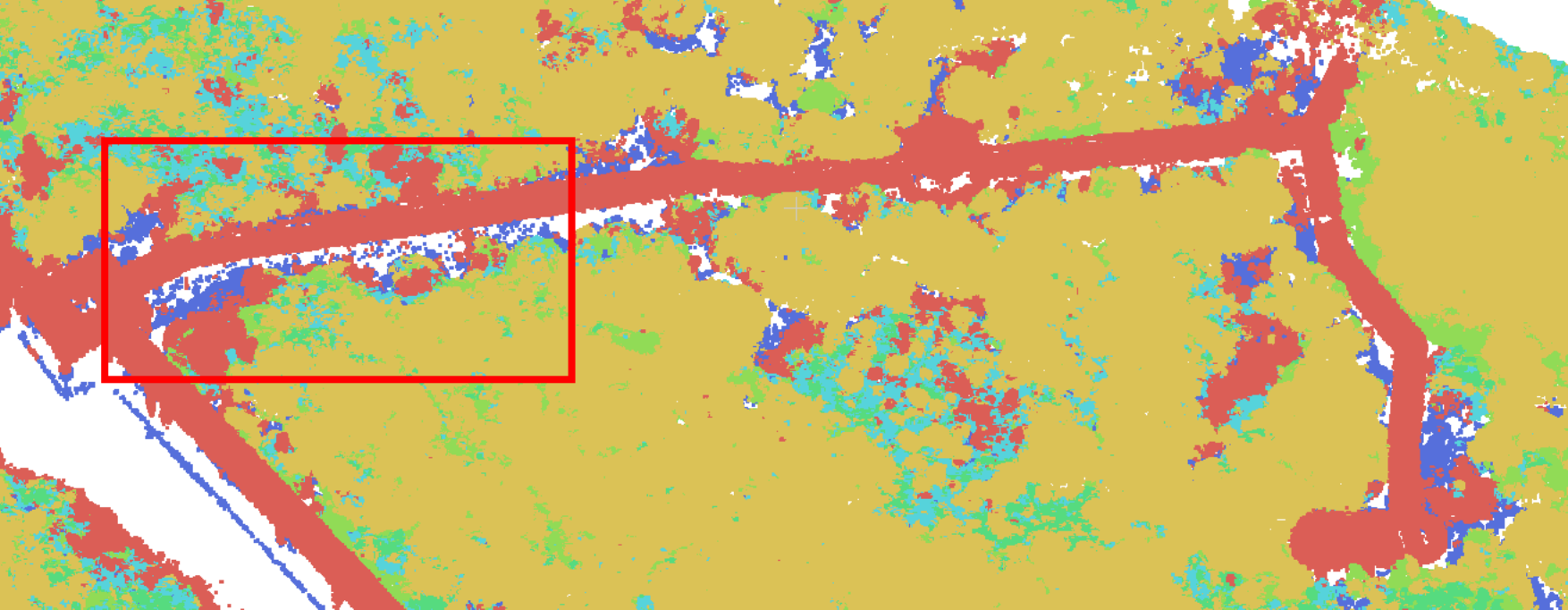}
            \end{minipage}
        }
    \end{minipage}
    \hspace{0.02\textwidth}
    \begin{minipage}[b]{0.30\textwidth}
        \centering
        \subfloat[]{
            \begin{minipage}[t]{0.95\textwidth}
                \includegraphics[width=\linewidth]{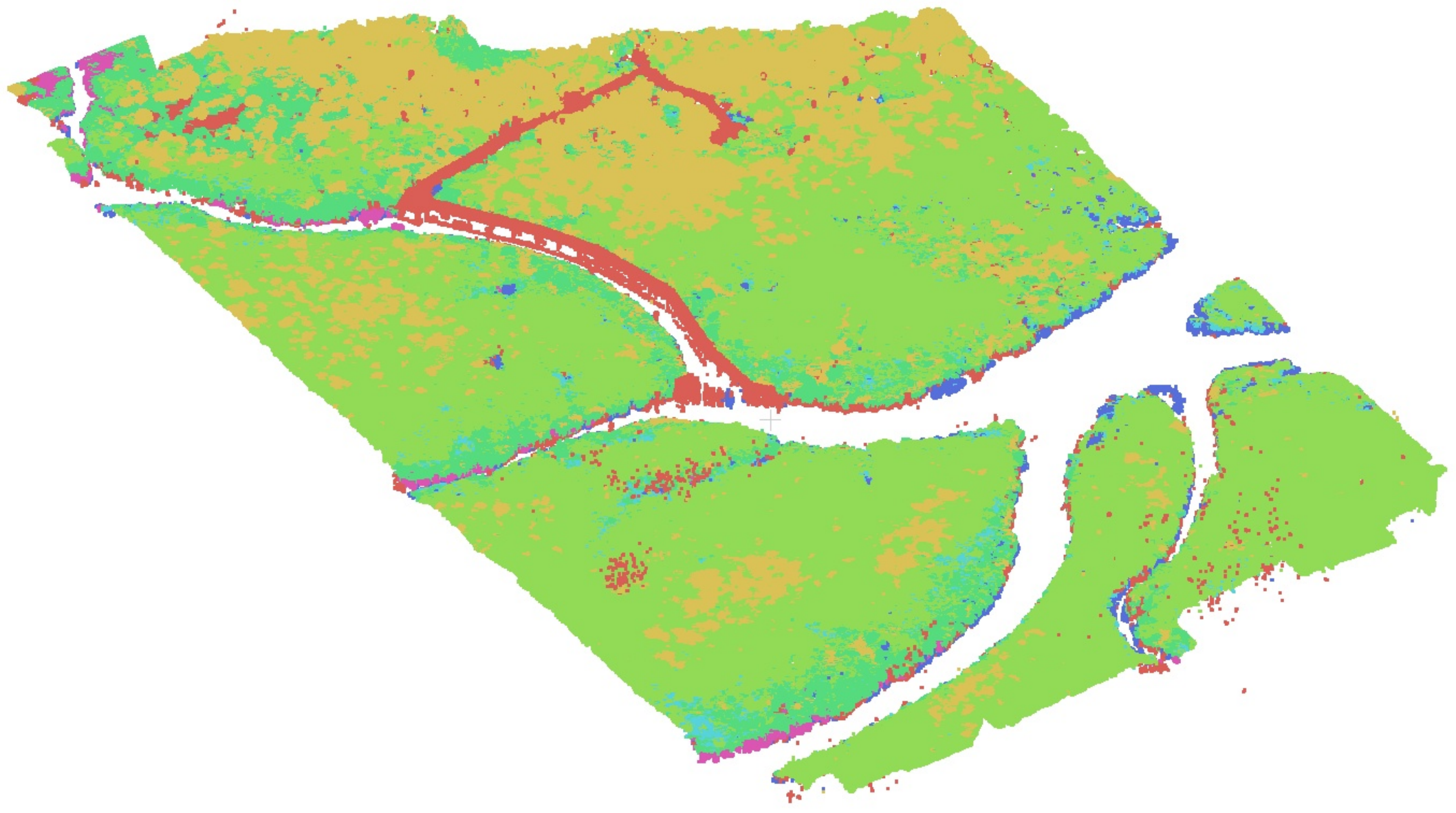}
                \includegraphics[width=\linewidth]{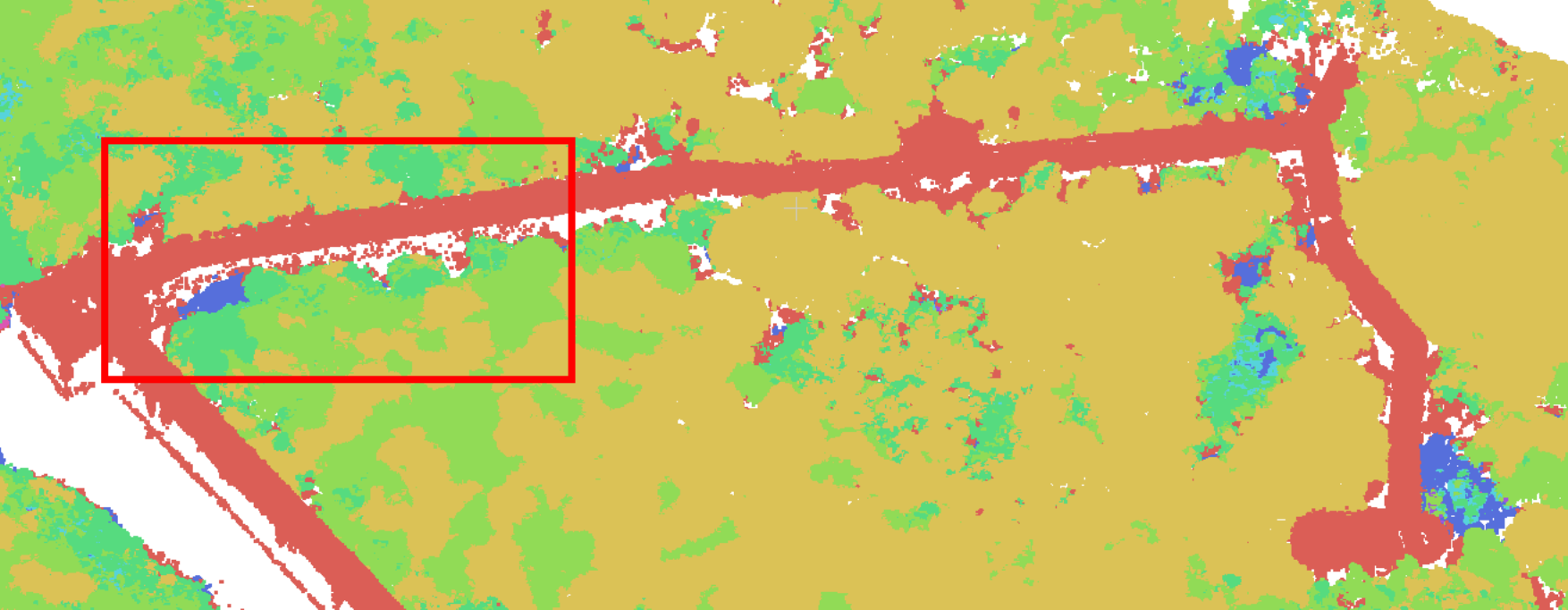}
            \end{minipage}
        }
    \end{minipage}
    \hspace{0.02\textwidth}
    \begin{minipage}[b]{0.30\textwidth}
        \centering
        \subfloat[]{
            \begin{minipage}[t]{0.95\textwidth}
                \includegraphics[width=\linewidth]{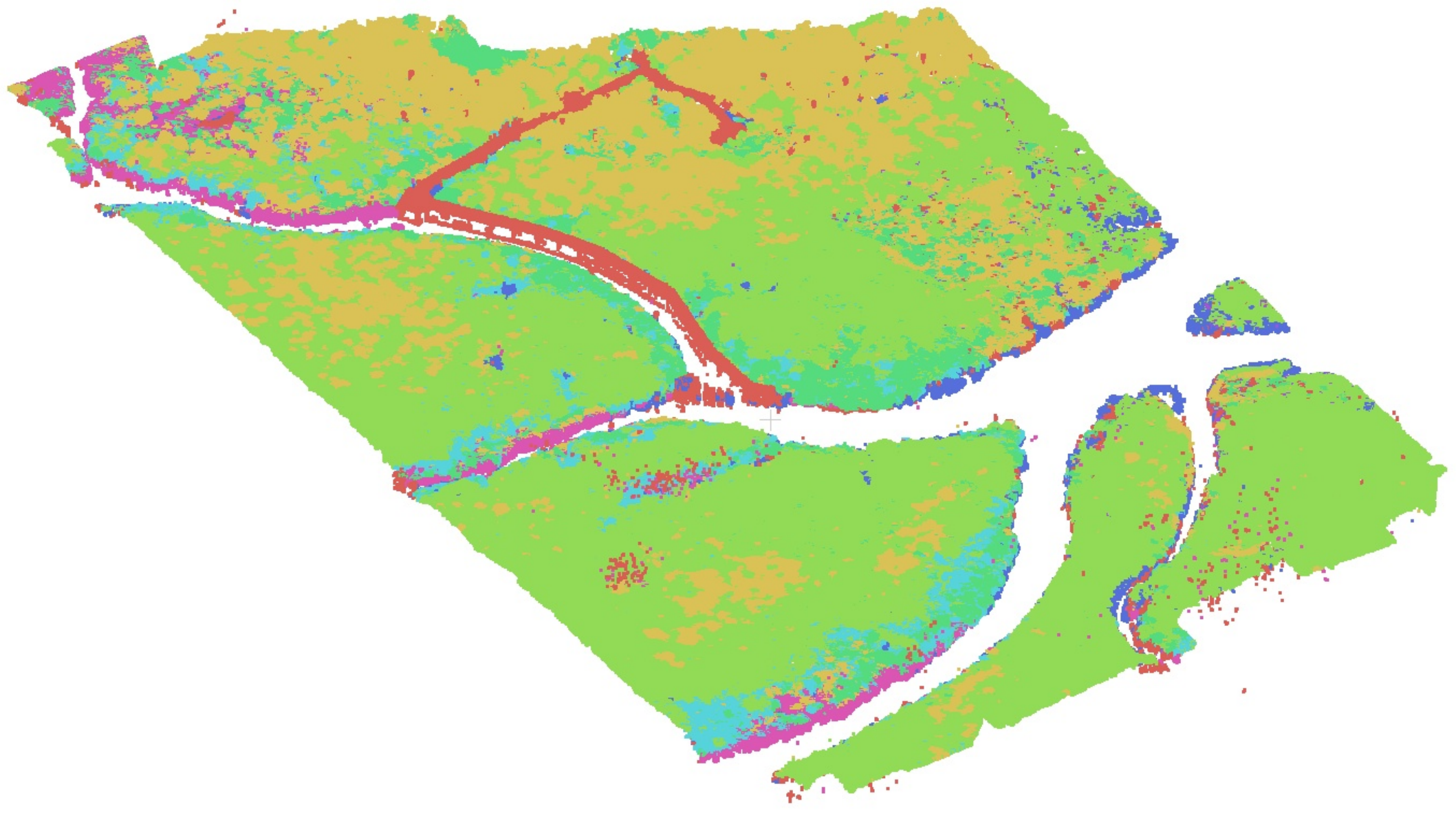}
                \includegraphics[width=\linewidth]{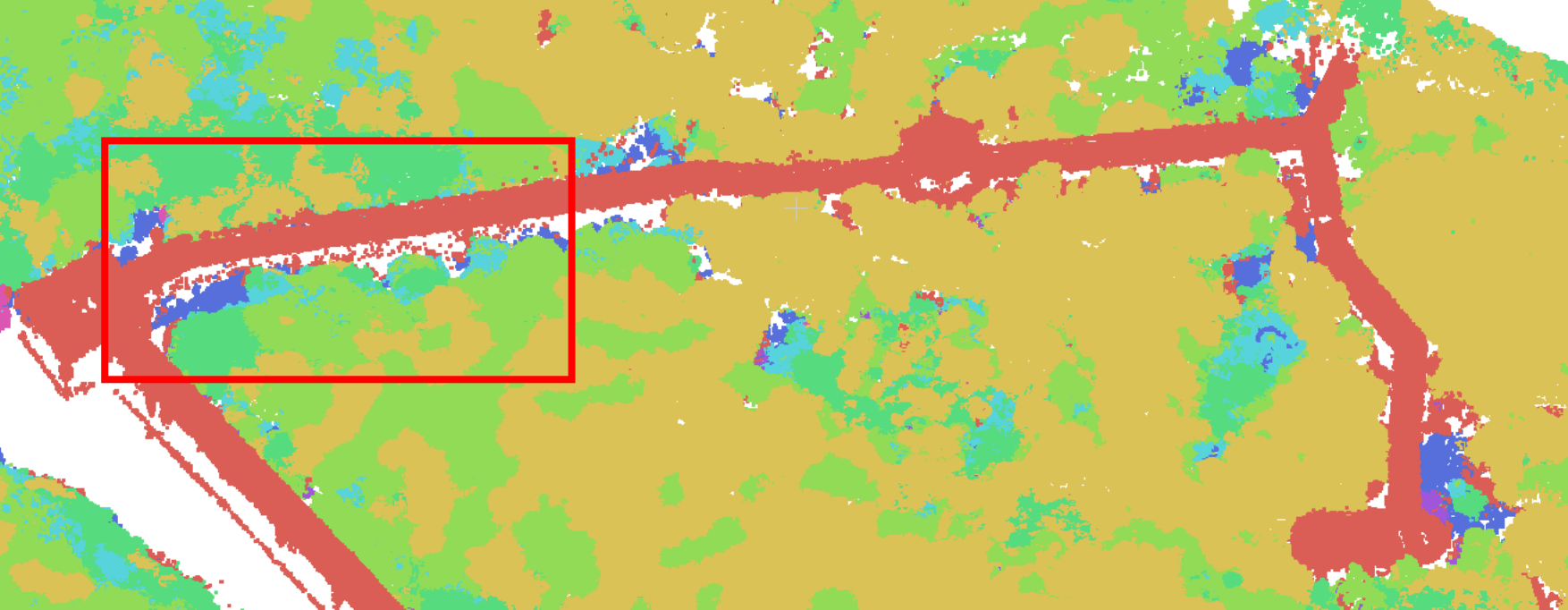}
            \end{minipage}
        }
    \end{minipage}
    \hspace{0.02\textwidth}
    \begin{minipage}[b]{0.30\textwidth}
        \centering
        \subfloat[]{
            \begin{minipage}[t]{0.95\textwidth}
                \includegraphics[width=\linewidth]{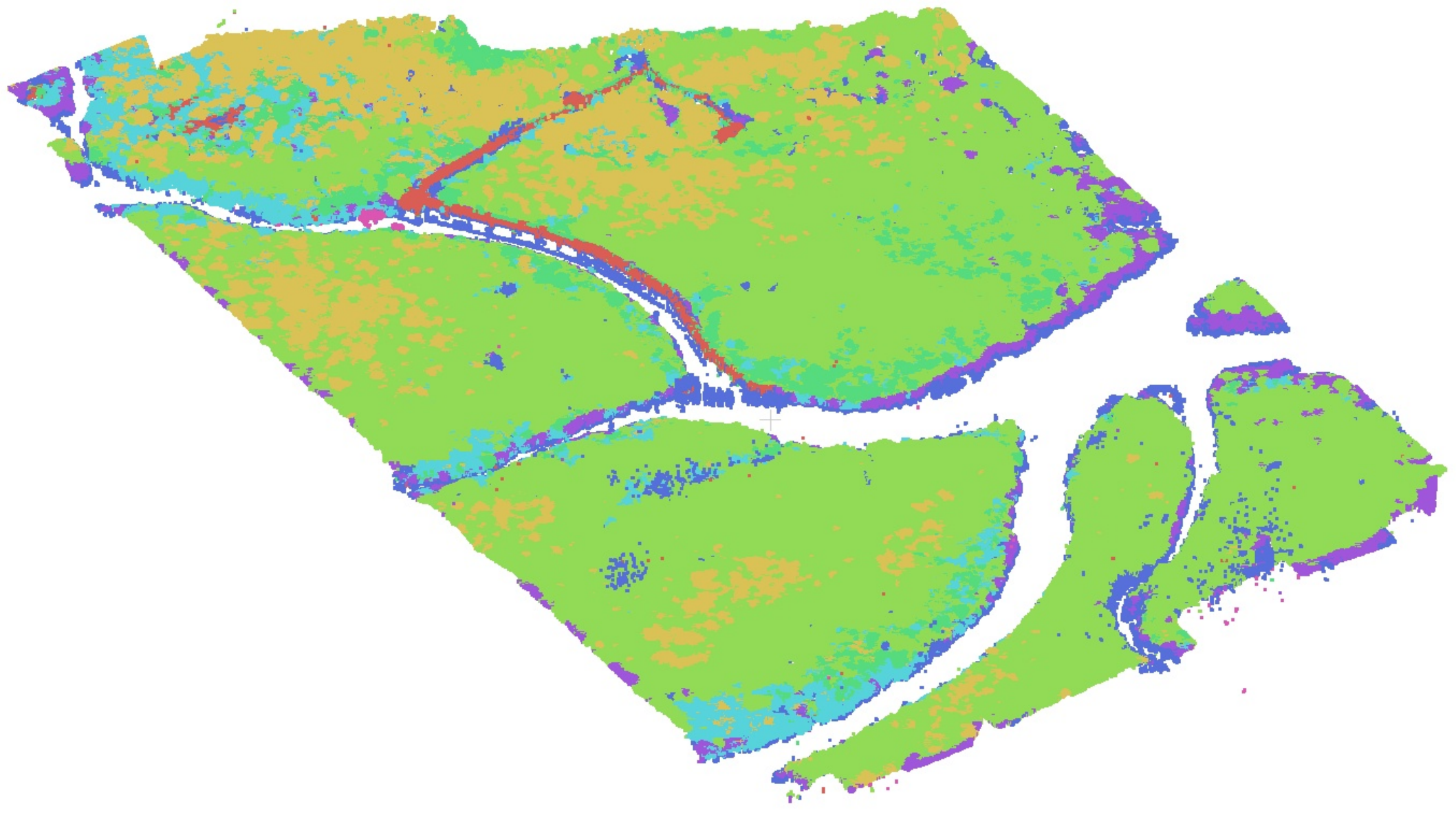}
                \includegraphics[width=\linewidth]{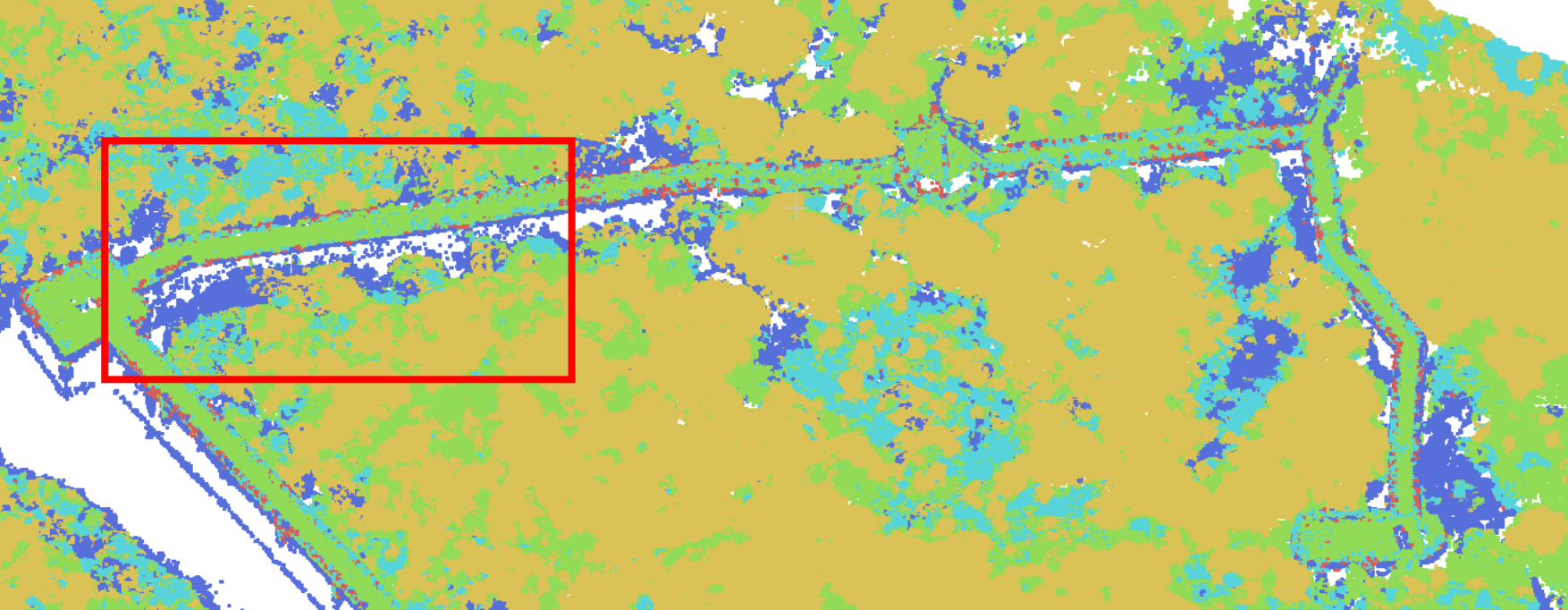}
            \end{minipage}
        }
    \end{minipage}
    \hspace{0.02\textwidth}
    \begin{minipage}[b]{0.30\textwidth}
        \centering
        \subfloat[]{
            \begin{minipage}[t]{0.95\textwidth}
                \includegraphics[width=\linewidth]{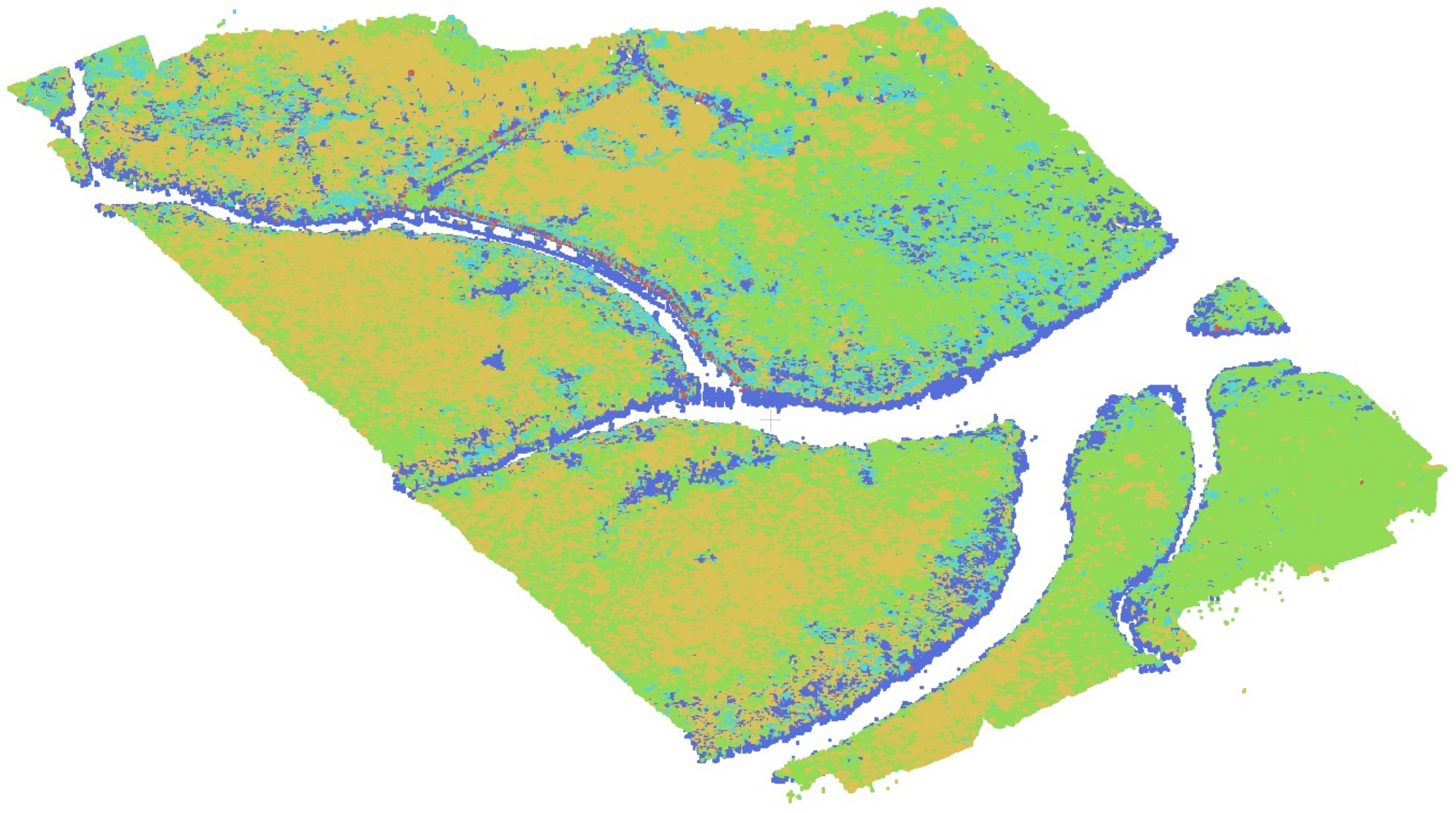}
                \includegraphics[width=\linewidth]{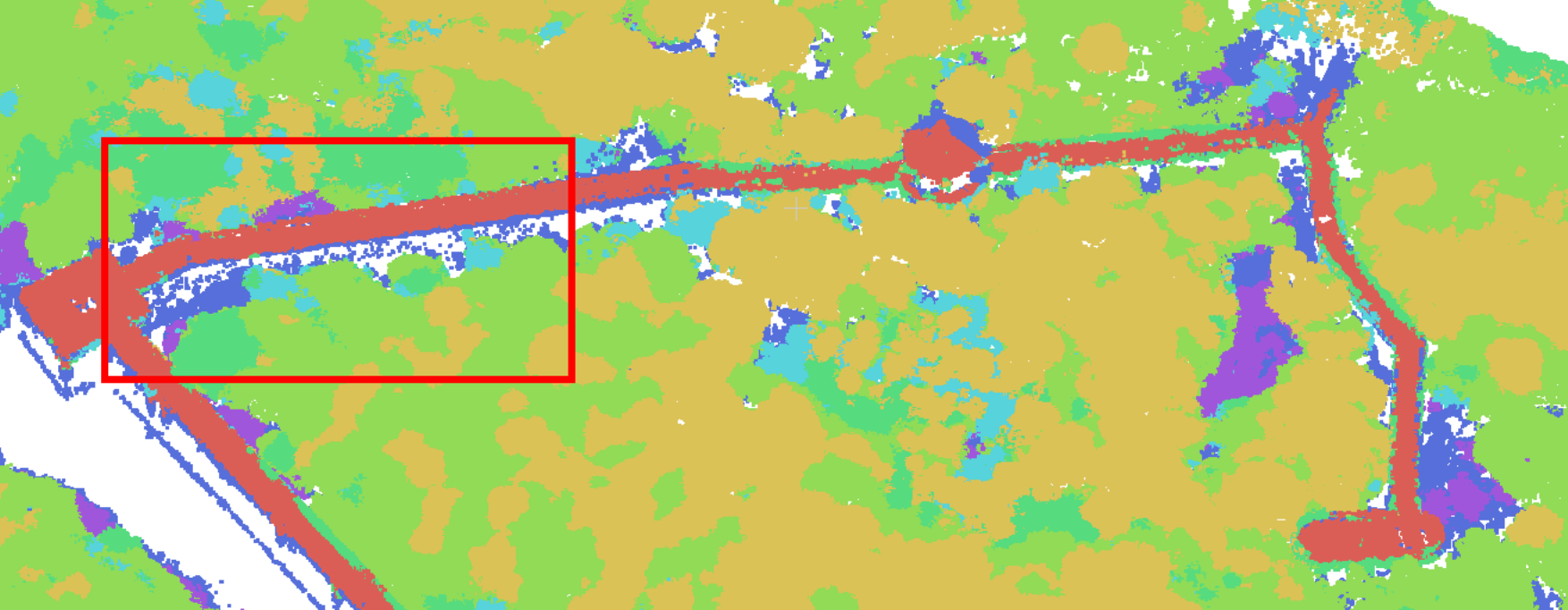}
            \end{minipage}
        }
    \end{minipage}
    \begin{minipage}[b]{0.98\textwidth}
        \centering
        \includegraphics[width=0.8\linewidth]{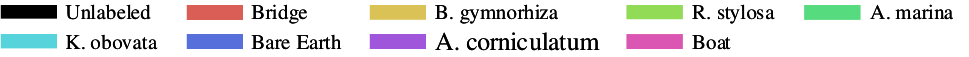}
    \end{minipage}
    \caption{Full classification maps on the SK dataset obtained by (a) Ground truth, (b) RandLA-Net, (c) SCFNet, (d) CGANet, (e) Hu et al, (f) Ours.}
    \label{fig8}
    \end{figure*}

Apart from quantitative analysis, Figure~\ref{fig6}, Figure~\ref{fig7} and Figure~\ref{fig8} show the full classification maps, as well as detailed maps of localized classifications. As shown in the red box, the proposed method can provide better classification results, such as the parking lot area in the HIT dataset in Fig. 6 and the area near the bridge in the SK dataset in Fig. 8. Also the proposed method is more consistent with the sparsely labeled ground truths, and for the unlabeled regions, the proposed method gives smoother and consistent prediction results, which are more in line with the land-covers distribution law in real scenes.

\subsection{Ablation study}
In this section, we verify the effectiveness of each module in the proposed network under different settings.

\subsubsection{Effectiveness of the proposed grid-balanced sampling}
First, we compare the proposed grid-balanced sampling (GBS) strategy with the traditional random sampling (RS) strategy. We changed the GBS strategy in the proposed network to the RS strategy adopted in~\cite{34} to minimize the overlap between training samples and test samples, then we ensure that the number of samples generated by the RS strategy is consistent with the GBS strategy, while other parameters remain unchanged. Comparing the difference between its results and the results of our proposed method. As shown in Table~\ref{tab4}, in the three datasets referenced in this study, higher overall indicators can be achieved when applying the GBS strategy, indicating that the GBS strategy is better at extracting classification features from sparsely labeled MPC compared to the RS strategy. In addition, since the RS strategy cannot guarantee to cover the complete test datasets, it requires more prediction time than the GBS strategy.

\begin{table}[!t]
    \centering
    \footnotesize
    \caption{Comparison of grid-balanced sampling strategy effects on three datasets.}
        \begin{tabular}{lcccccc}
        \toprule
        \multirow{2}[2]{*}{} & \multicolumn{2}{c}{HIT dataset} & \multicolumn{2}{c}{ZJK dataset} & \multicolumn{2}{c}{SK dataset} \\
                & RS & GBS & RS & GBS & RS & GBS \\
        \midrule
        OA & 76.21 & \textbf{87.07} & 68.44 & \textbf{91.73} & 36.8  & \textbf{86.91} \\
        AA & 70.07 & \textbf{90.2} & 24.88 & \textbf{78.96} & 27.54 & \textbf{83.59} \\
        kappa & 72.14 & \textbf{84.89} & 29.6  & \textbf{85.82} & 24.54 & \textbf{82.03} \\
        mIoU & 49.58 & \textbf{67.66} & 16.75 & \textbf{66.93} & 10.75 & \textbf{67.8} \\
        \midrule
        Time & 1738 & \textbf{1241} & \textbf{1608} & 1629 & 1949  & \textbf{1525} \\
        \bottomrule
        \end{tabular}
    \label{tab4}
\end{table}

\subsubsection{Analysis of different hyperparameters}
\begin{figure*}[!t]
    \centering
    \begin{minipage}[b]{0.30\textwidth}
        \centering
        \subfloat[]{
            \begin{minipage}[t]{0.95\textwidth}
                \includegraphics[width=\linewidth]{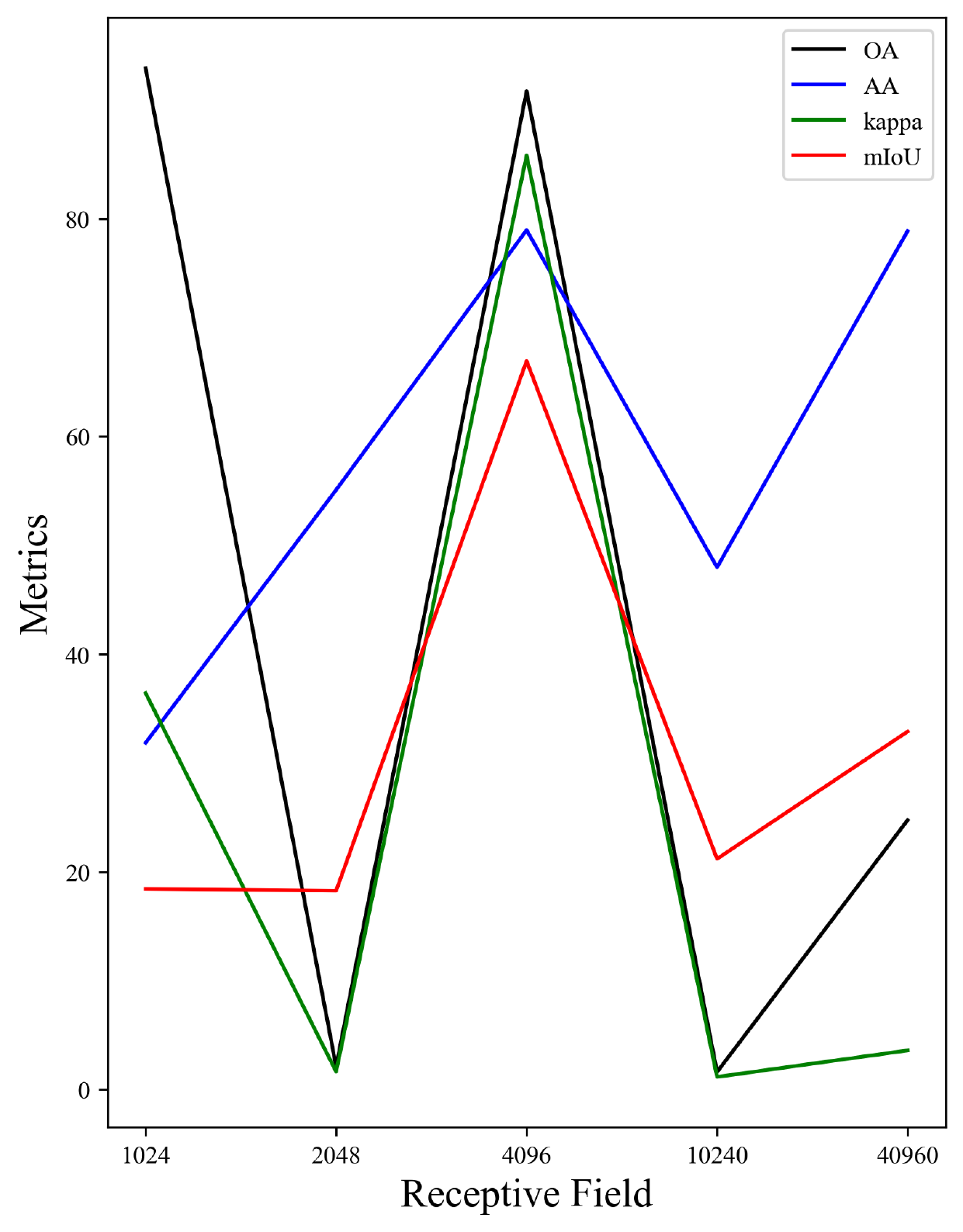}
            \end{minipage}
    }
    \end{minipage}
    \centering
    \begin{minipage}[b]{0.30\textwidth}
        \centering
        \subfloat[]{
            \begin{minipage}[t]{0.95\textwidth}
                \includegraphics[width=\linewidth]{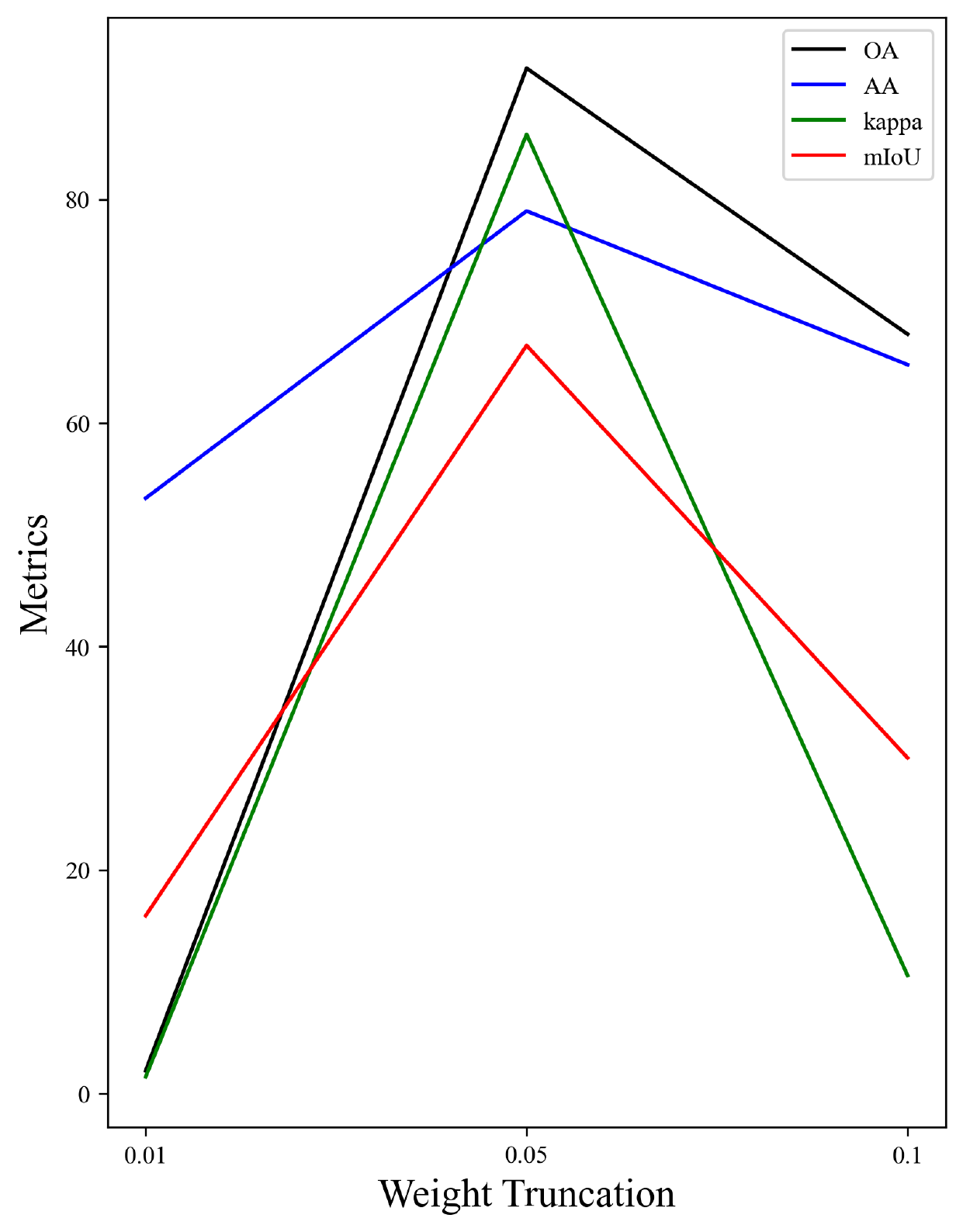}
            \end{minipage}
    }
    \end{minipage}
    \centering
    \begin{minipage}[b]{0.30\textwidth}
        \centering
        \subfloat[]{
            \begin{minipage}[t]{0.95\textwidth}
                \includegraphics[width=\linewidth]{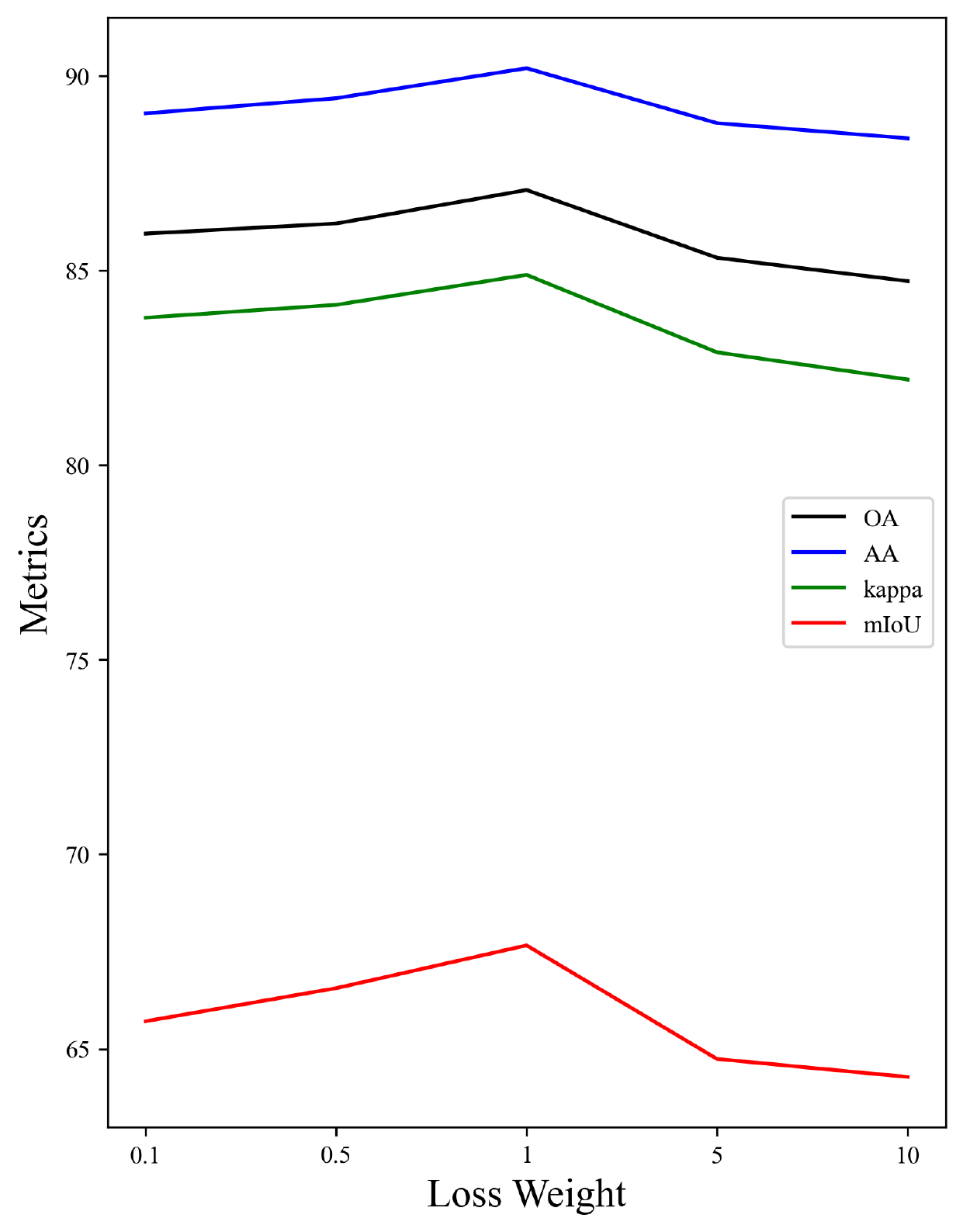}
            \end{minipage}
    }
    \end{minipage}
    \caption{Comparison of different hyperparameters. (a) Comparison of different receptive field. (b) Comparison of different weight truncation. (c) Comparison of different loss weight.}
    \label{fig9}
    \end{figure*}

(1) Receptive field. To validate the effectiveness of the proposed receptive field sizes, five sets of experiments with different receptive fields are set: 1024, 2048, 4096, 10240, and 40960. The effect of the above five receptive field sizes is tested on the proposed method on the ZJK datasets, respectively. The results in Figure~\ref{fig9} (a) show that the highest overall accuracy can be achieved when the size of the receptive field is 4096. This is because when the receptive field is too small, land-covers with large scales (e.g., buildings) can only learn some parts of the features, which leads to a decrease in the classification accuracy; When the receptive field is too large, due to the sparse labeled of ground truths, there are a large number of unlabeled points in the receptive field, which reduces the learning efficiency of the network. Therefore the results are in line with the theoretical expectations.

(2) Weight truncation. The ZJK dataset's feature distribution is too uneven, with the tidal flat category accounting for about 95\% of the total points. Therefore, when the learning rate of different categories is normalized according to the proportion of each category, it is easy to cause the learning rate of the tidal flat category to disappear, resulting in the inability to learn effective features. three sets of experiments paired with different learning rate weights truncation are set: 0.01, 0.05, 0.10. The results on the ZJK dataset shown in Figure~\ref{fig9} (b) indicate that the highest overall accuracy can be achieved when the weight truncation is 0.05.

(3) Loss weight. The number of loss weight $\lambda$ is a key hyperparameter in our method, which determines the ability of our network to balance the ability to learn multi-scale with long-tailed land covers. Apparently, a larger $\lambda$ strengthens the network's ability to learn multi-scale features and correspondingly weakens the network's ability to learn long-tailed features. Five sets of experiments with different loss weights are set: 0.1, 0.5, 1, 5, and 10. The effect of the above five loss weights is tested on the proposed method on the HIT datasets, respectively. The results in Figure~\ref{fig9} (c) show that the highest overall accuracy can be achieved when the loss weight is 1.

\subsubsection{Effectiveness of the proposed module}
To validate the importance of each module in the network, ablation experiments are conducted using the HIT dataset. The details of ablation studies and evaluation results are presented in the following.

We still use the RandLA-Net model as the backbone network and apply the MSFF and AHL modules proposed in this paper to compare the results of each model on the HIT dataset separately. As seen in Table~\ref{tab5}, five sets of control experiments are designed, they are 1) backbone network; 2) adding the MSFF module (+MSFF); 3) adding the multi-scale loss module (+MSL); 4) adding the long-tailed loss module (+LTL); 5) adding the AHL module (+AHL); 6) adding the MSFF and AHL module (Ours).

\begin{table}[!t]
    \centering
    \footnotesize
    \caption{Ablation resuls for different modules on HIT dataset. 1) backbone network; 2) adding the MSFF module (+MSFF); 3) adding the multi-scale loss module (+MSL); 4) adding the long-tailed loss module (+LTL); 5) adding the AHL module (+AHL); 6) adding the MSFF and AHL module (Ours).}
        \begin{tabular}{lcccccc}
        \toprule
        Model & Baseline & +MSFF & +MSL & +LTL  & +AHL  & \textbf{Ours} \\
        \midrule
        MSFF & & \checkmark & & & & \checkmark \\
        MSL & & & \checkmark & & \checkmark & \checkmark \\
        LTL & & & & \checkmark & \checkmark & \checkmark \\
        OA & 78.36 & 84.61 & 84.90 & 85.28 & 85.99 & \textbf{87.07} \\
        \textbf{$\Delta$ OA} & 0 & +6.25 & +6.54 & +6.92 & +7.63 & \textbf{+8.71} \\
        AA & 84.48 & 88.03 & 87.16 & 88.29 & 88.28 & \textbf{90.2} \\
        \textbf{$\Delta$ AA} & 0 & +3.55 & +2.68 & +3.81 & +3.80 & \textbf{+5.72} \\
        kappa & 74.97 & 82.07 & 82.45 & 82.83 & 83.61 & \textbf{84.89} \\
        \textbf{$\Delta$ kappa} & 0 & +7.10 & +7.48 & +7.86 & +8.64 & \textbf{+9.92} \\
        mIoU & 61.41 & 67.65 & 67.15 & 67.05 & 67.16 & \textbf{67.66} \\
        \textbf{$\Delta$ mIoU} & 0 & +6.24 & +5.74 & +5.64 & +5.75 & \textbf{+6.25} \\
        head avg. & 75.91 & 82.64 & 83.89 & 83.17 & 83.47 & \textbf{84.56} \\
        \textbf{$\Delta$ head avg.} & 0 & +6.73 & +7.98 & +7.26 & +7.56 & \textbf{+8.65} \\
        tail avg. & 90.61 & 91.88 & 89.49 & 91.94 & 91.72 & \textbf{94.22} \\
        \textbf{$\Delta$ tail avg.} & 0 & +1.27 & -1.12 & +1.33 & +1.11 & \textbf{+3.61} \\
        head min. & 36.97 & 57.46 & \textbf{70.24} & 66.35 & 60.97 & 64.72 \\
        \textbf{$\Delta$ head min.} & 0 & +20.49 & \textbf{+33.27} & +29.38 & +24.00 & +27.75 \\
        tail min. & 72.17 & 77.89 & 63.18 & 75.70 & 72.93 & \textbf{87.06} \\
        \textbf{$\Delta$ tail min.} & 0 & +5.72 & -8.99 & +3.53 & +0.76 & \textbf{+14.89} \\
        \bottomrule
        \end{tabular}
    \label{tab5}
\end{table}

The results of the experiments are shown in Table~\ref{tab5}. It can be seen that the above five groups of experiments can achieve OA enhancement of 6.25\%, 6.54\%, 6.92\%, 7.63\%, and 8.71\%, respectively. Among them, the MSFF and MSL modules provide the largest improvement in classification accuracy for the head category, especially for head min. which provides 20.49\% and 33.27\% improvement respectively. This is consistent with the observation that large-scale land-covers generally accounts for a larger portion of the points as well, while the LTL module provides a larger boost for the tail category, consistent with the module objective. When all the above modules are applied simultaneously, it can be observed that the proposed method achieves the highest overall metrics, suggesting that a better combination between the different modules is reached.

\section{Conclusion}
In this work, an enhanced classification method based on adaptive multi-scale fusion for MPCs with long-tailed distributions was proposed. A novel training set generation method based on the grid-balanced sampling strategy was designed for robust training set extraction from sparsely labeled MPC. Subsequently, a multi-scale feature fusion module was developed for enhancing the preservation of fine features at different scales in the feature aggregation network. Finally, an adaptive hybrid loss method that contains a multi-scale loss, and long-tailed loss was devised to address the scale difference problem and the long-tailed distribution problem of land-covers in outdoor datasets, respectively. Experimental results on three datasets illustrated the effectiveness of the proposed network and modules.

Although the proposed method achieves remarkable accuracy in classification for MPCs, one limitation is that the grid-balanced sampling strategy is prone to the one-shot problen when facing densely distributed categories, leading to overfitting. For future work, we will further explore advanced techniques to alleviate the one-shot issue.

\section{Acknowledgements}
This work was supported in part by the National Natural Science Foundation of China through the Youth Science Foundation Project under Grant 62301192 and Outstanding Young Scholars under Grant 62025107, and in part by the Postdoctoral Science Foundation under Grant Number 2024T171158.

\bibliographystyle{unsrtnat}


\end{document}